\documentclass[10pt,twocolumn,letterpaper]{article}

\usepackage{cvpr}

\usepackage{booktabs}
\usepackage{colortbl}

\definecolor{qcSand}{RGB}{252,246,232}
\definecolor{qcSandBest}{RGB}{244,232,204}
\definecolor{qcTeal}{RGB}{226,239,236}
\definecolor{qcTealBest}{RGB}{198,224,216}
\definecolor{qcSlate}{RGB}{232,237,242}
\definecolor{qcSlateBest}{RGB}{210,221,232}
\definecolor{qcGold}{RGB}{243,236,214}
\definecolor{qcClay}{RGB}{246,236,230}
\usepackage{amsmath,amssymb}
\usepackage{graphicx}
\usepackage{tikz}
\usetikzlibrary{arrows.meta,positioning,fit}

\usepackage{microtype}

\makeatletter
\def\cvprsection{\@startsection {section}{1}{\z@}
   {-8pt plus -1pt minus -1pt}{5pt} {\large\bf}}
\def\cvprsubsection{\@startsection {subsection}{2}{\z@}
   {-6pt plus -1pt minus -1pt}{3pt} {\elvbf}}
\makeatother
\renewcommand{\paragraph}[1]{\vspace{0.2em}\noindent{\tenbf #1}\hspace{0.35em}}
\definecolor{cvprblue}{rgb}{0.21,0.49,0.74}
\usepackage[pagebackref,breaklinks,colorlinks,allcolors=cvprblue]{hyperref}

\def\confName{CVPR}
\def\confYear{2027}

\makeatletter
\def\@maketitle{%
  \newpage
  \null
  \vskip 0.12in
  \begin{center}%
    {\LARGE\bfseries \@title \par}%
    \vskip 16pt
    {\large \@author \par}%
  \end{center}%
  \vskip 6pt
}
\makeatother

\title{PhoneQC: Benchmarking Multimodal LLMs for Order-Level Smartphone Trade-in Inspection}

\author{%
Zihan Deng$^{1}$ \quad Xinpei Lin$^{2}$ \quad Du Tiantian$^{1}$ \quad Lequan Yu$^{1}$\\[0.7em]
{\normalsize $^{1}$The University of Hong Kong \quad $^{2}$Nanchang University}%
}

\begin{document}
\maketitle
\begin{abstract}
As vision-language models enter remote smartphone trade-in, sellers already upload six views and a diagnostic screen, which makes it easy to treat the human quote as a step a model can take over.
A model may mark a crack or read the screen, yet the quote is a staff form: appearance fill stays near 0.25, and the closed models that score a little higher do so only on the orders they answer, so human appraisal remains necessary.
Two obstacles stand in the way.
(1) Single-view defect sets and attribute benchmarks do not score an order-level form under a fixed input contract, so a box or a leaked field can be mistaken for a finished quote.
(2) The unread half of the form is the staff fields, and a model left to emit boxes or captions does not learn them.
We present PhoneQC, a 335-order iPhone-heavy cut that ranks systems by a 0 to 100 form score (Score100) only inside matched protocols (images alone; images plus a detector sidecar), with a declared upper bound that is not ranked as a peer.
As a case study on that board, field-only LoRA freezes the backbone, adapts attention projections, and supervises staff fields while leaving the defect list empty; the adapter sees the same images as the zero-shot model.
On image-only T1 the 3B model rises from 26.6 to 33.3 at coverage 0.928; without the empty-blocker item the open ranking still favors the adapter, while the remaining gap to a person stays large on this cut.
\end{abstract}

\section{Introduction}

\begin{figure}[t]
\centering
\includegraphics[width=\linewidth]{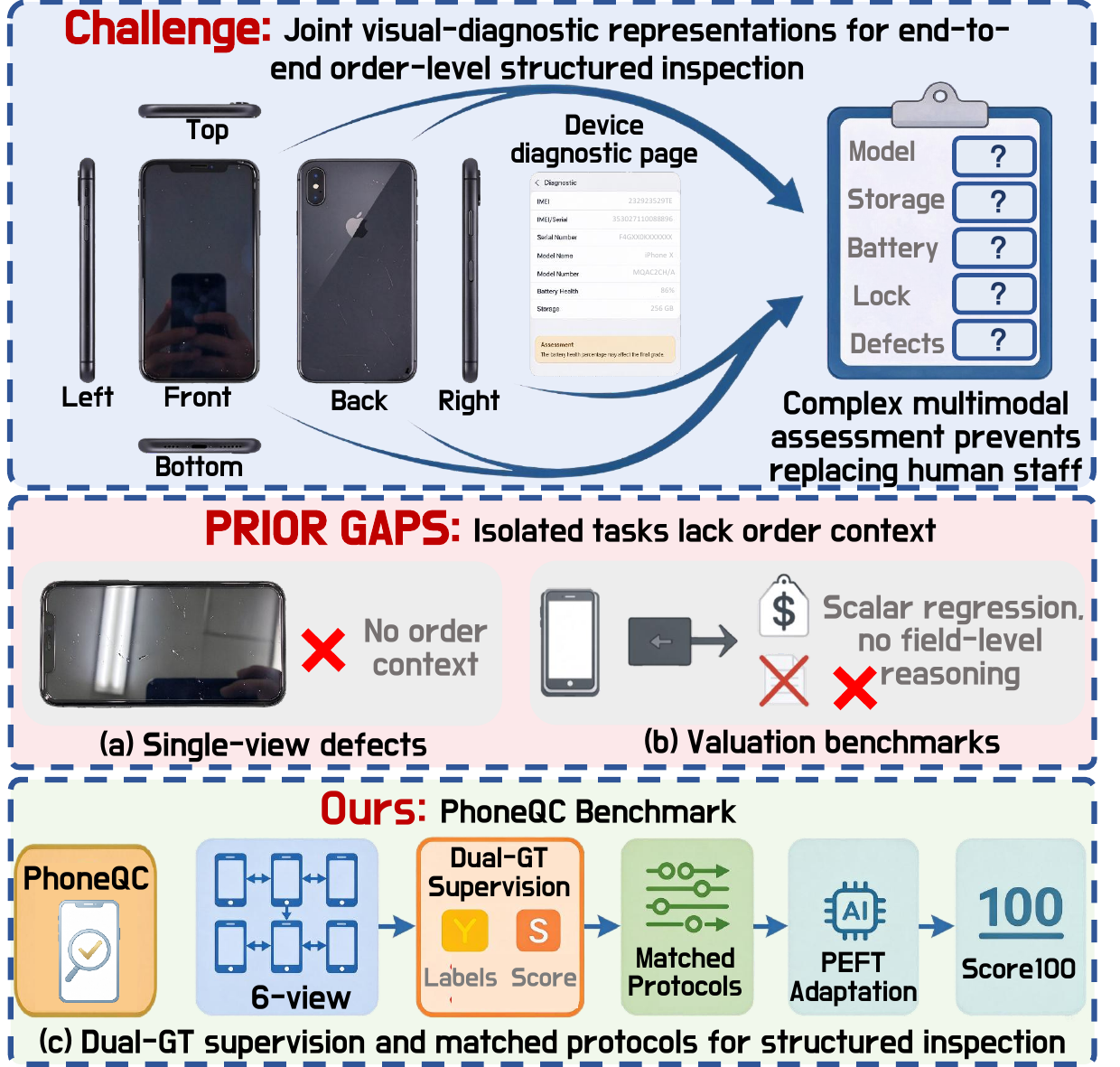}
\caption{PhoneQC asks for an order-level form, not a price.
Six exterior views and a diagnostic screen must fill model, storage, battery, lock, and defects.
(a)~Single-view defect data has no order.
(b)~Valuation benchmarks reduce the job to a scalar and skip field-level reasoning.
(c)~PhoneQC pairs dual supervision (appearance items $Y$ and staff fields $S$) with matched protocols, low-rank adaptation (LoRA), and Score100.}
\label{fig:motivation}
\end{figure}

Remote smartphone trade-in inspection must be completed before the buyer receives the device. Sellers submit an order containing exterior photographs and, when available, a diagnostic screen with model, storage, and battery information. Professional appraisers review this evidence and fill a staff form covering device identity, condition, lock status, blockers, and defects. A wrong form field can change the repurchase offer even when a crack is localized correctly. Vision-language models make automated inspection plausible, but the operational output is a structured staff form rather than a price or an isolated defect box. Progress therefore requires an evaluation protocol that specifies the permitted inputs and scores the complete form.

Factory defect sets already localize cracks with care.
The Smartphone Screen Glass Dataset (SSGD) photographs smartphone glass on a fixed workstation, while MVTec AD, the visual anomaly set VisA, and Kaputt do the same for industrial parts under controlled capture~\cite{ssgd2023,mvtec2019,visa2022,kaputt2025}.
M5Product, SMARTAVE (structured multimodal attribute values), and ImplicitAVE (implicit attribute values) recover catalog fields from product images~\cite{dong2022m5product,wang2022smartave,zou2024implicitave}, while DocVQA (document visual question answering), ScreenQA (screen question answering), MMBench, and MMMU (Massive Multi-discipline Multimodal Understanding) read interfaces or test open-domain competence~\cite{docvqa2021,screenqa2025,mmbench2024,mmmu2024}.
A correct box on one view still leaves storage, lock, and battery unread, and none of these resources scores the views and the diagnostic screen as one staff form under a contract that states which images a system may see.

Together, these benchmarks leave two gaps: they do not score a complete order-level form under a declared input contract, and their usual box or caption targets do not train a model to fill staff fields.

We present PhoneQC, an order-level benchmark on a 335-order iPhone-heavy cut, with field-only LoRA as a case study (Figure~\ref{fig:motivation}). Each protocol declares its inputs, keeping detector sidecars and staff labels out of the image-only evaluation. We freeze Qwen2.5-VL and adapt its attention projections to supervise staff fields, leaving the defect list empty~\cite{qwen25vl2025}. At test time, every system in one protocol receives the same deterministic image pack. On T1, the 3B model's Score100 rises from 26.6 to 33.3 at coverage 0.928, while appearance fill remains near one quarter. Since the reference blocker list is empty for every order, S10 rewards empty predictions; we therefore also report Score100 without S10 in the supplement. A detector sidecar raises box F1 under T2 in these logs, but does not consistently raise Score100. Our contributions are as follows.

PhoneQC Benchmark.
We construct a 335-order form-score cut with appearance labels and staff-form references, Score100, and matched input protocols that keep side information off the image-only rank.

Field-only LoRA case study.
We adapt attention projections of a frozen backbone under staff-field supervision with an empty defect list, and test the adapter on the same images as the zero-shot model among the schedules logged here.

Evaluation practice.
We report Score100 with coverage, disclose the empty-blocker item, and expand no-S10 and common-answered checks in the supplement; results are observations on this cut, not a claim that appraisal is replaced.

\section{Related Work}

\subsection{Single-view defects and attributes}
Work closest to trade-in inspection falls into two groups: methods that localize defects, and methods that recover product attributes.
Neither group evaluates an order-level staff form under a fixed input contract.

Factory and logistics anomaly sets study cracks, stains, and missing parts under controlled capture.
MVTec AD~\cite{mvtec2019}, VisA~\cite{visa2022}, and Kaputt~\cite{kaputt2025} provide boxes or masks on industrial or package views, and SSGD photographs smartphone glass on a fixed workstation and scores that one pane~\cite{ssgd2023}.
These resources are strong on a single view, but they leave the other sides of a consumer upload, the diagnostic screen, and the staff fields unread.

Attribute extraction moves closer to catalog text.
M5Product pretrains on e-commerce images and text~\cite{dong2022m5product}, and SMARTAVE and ImplicitAVE recover structured or implied attribute values from product images~\cite{wang2022smartave,zou2024implicitave}.
That line answers which fields appear in a listing, yet it does not state which pixels a system may see before the answer is scored, and it does not treat the full inspection form as the rank.
What remains missing is an order scored as one staff form under a contract that keeps privileged inputs off the image-only board.

\subsection{Vision-language models}
Vision-language models supply the other half of the stack.
CLIP aligns an image with a text description~\cite{radford2021clip}.
Flamingo and BLIP-2 attach a frozen visual encoder to a language model~\cite{alayrac2022flamingo,li2023blip2}.
LLaVA, the later LLaVA-1.5 baseline, and InstructBLIP instruction-tune that interface so the model emits an answer~\cite{liu2023llava,liu2024llava15,dai2023instructblip}.
Those answers are usually scored as visual questions, document or screen reading, or a broad multimodal suite~\cite{antol2015vqa,docvqa2021,screenqa2025,mmbench2024,mmmu2024,li2024seed}, while Rico and Android in the Wild target mobile interfaces and device control~\cite{rico2017,aitw2023}.

The limiting factor is the unit of evaluation.
A multiple-choice item, a document page, or a single screenshot is not a six-view trade-in order: those benchmarks do not declare which extra fields are forbidden at test time, and they do not average a gated form score over an answered set while reporting coverage.
Open adapters usually supervise a caption, a box list, or a free-form reply, which does not force the model to write the staff half of an inspection form while leaving the defect list empty.
The next sections fix that unit of evaluation and train an adapter on the staff fields alone.

\section{The PhoneQC Benchmark}
\label{sec:protocol}

\begin{figure*}[t]
\centering
\includegraphics[width=\textwidth]{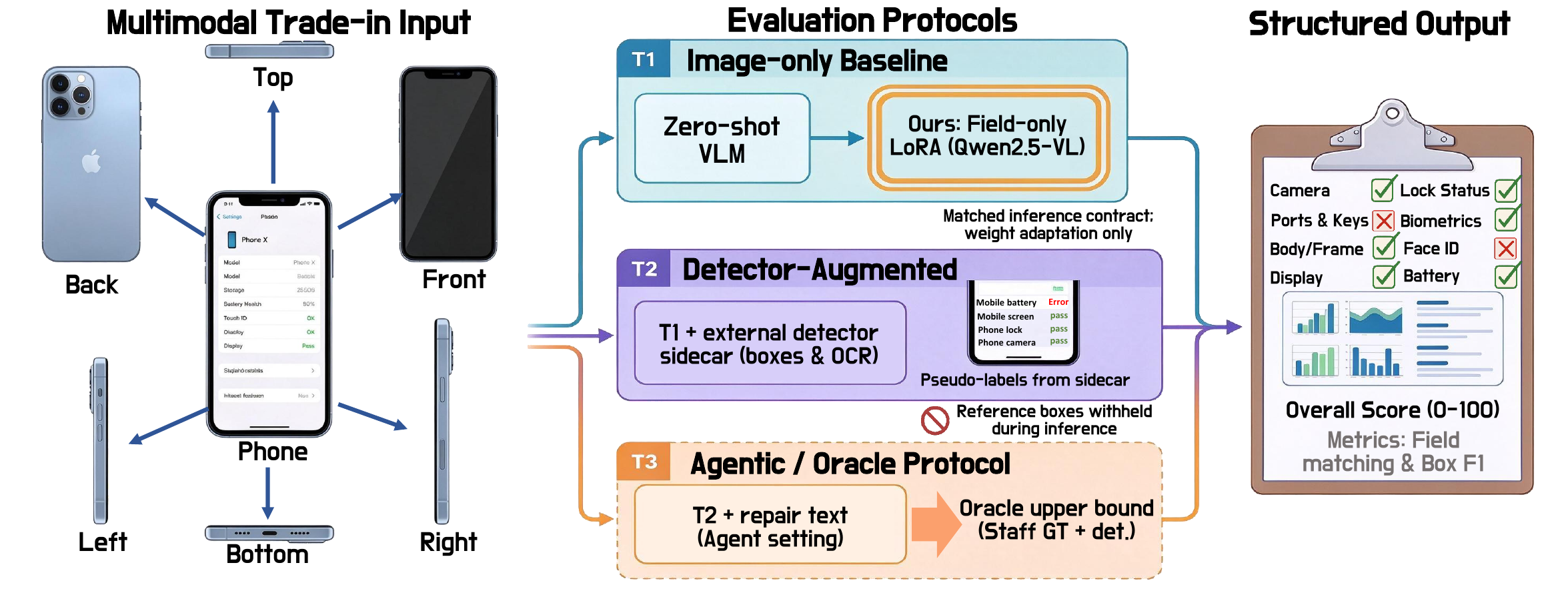}
\caption{Matched protocols for one order.
The input is six exterior views plus the diagnostic screen.
T1 is image-only and contains both a zero-shot vision-language model (VLM) and our field-only LoRA. Adaptation changes weights, and the images stay the same.
T2 adds an external detector sidecar, with boxes and a short optical character recognition (OCR) snippet.
Reference boxes stay hidden.
T3 may add declared repair text. The oracle that also reads staff fields is an upper bound, and it is not a T1 result.
Every rail emits the same form, scored by field matching and a box F1 score.}
\label{fig:method}
\end{figure*}

PhoneQC treats each trade-in order as one evaluation unit whose answer is a staff form, not a repurchase price and not a single defect box.
Before the score is computed, every comparison must state which photographs a system may see and which privileged fields remain hidden, so that a detector sidecar or a leaked staff label cannot enter the image-only rank by accident.

\paragraph{Orders and the staff form.}
Each test item is one trade-in order whose visible evidence is the six exterior views when the seller provided them, plus a diagnostic screenshot when one was uploaded, and whose reference answer is the staff form written before the device arrives.
We write one order as $(X, a^\star)$, where $X$ is that upload and $a^\star$ is the staff form, and a system returns a decision $d\in\{\mathrm{answer},\mathrm{abstain},\mathrm{human\_review}\}$ together with a form $a$ in the same schema.
Let $\mathcal{A}$ be the set of orders whose decision is answer, and let $N$ be the size of the test cut.
Coverage and the official score are then
\begin{equation}
\mathrm{Cov}=\frac{|\mathcal{A}|}{N},\qquad
\overline{\mathrm{Score100}}=\frac{1}{|\mathcal{A}|}\sum_{i\in\mathcal{A}}\mathrm{Score100}(a_i,a_i^\star).
\label{eq:official}
\end{equation}
An order outside $\mathcal{A}$ changes coverage and leaves the answered mean untouched, so a model can look strong on $\overline{\mathrm{Score100}}$ simply by answering the orders it finds easy.
The leaderboard therefore reports both numbers, and a quoted Score100 is meaningful only when it carries the coverage, or the answered count, of that same row.
In particular, comparing a closed model that answers roughly three quarters of the cut with an open adapter that answers nearly all of it is a comparison of two different answered sets unless both numbers are shown.

\paragraph{Matched protocols.}
Figure~\ref{fig:method} fixes the input contract that every system must obey.
All three rails emit one form and differ only in the side information allowed to enter the prompt: T1 conditions on $X$ alone, T2 conditions on $X\cup D$ where $D$ is a detector sidecar of predicted boxes and a short screen transcript that must be disjoint from the reference boxes inside $a^\star$, and T3 may also read seller repair text.
The oracle additionally reads $a^\star$ itself and is scored as a ceiling under privileged inputs, drawn dashed in Figure~\ref{fig:method}, so it is not ranked as a peer of the image-only systems.
A comparison is valid only inside one of these inputs, and a gain that appears only after $D$ or $a^\star$ is added cannot be entered on the image-only board.
Zero-shot models and field-only LoRA share T1: the adapter changes weights and then sees the same $X$.
T2 asks whether $D$ helps the form while $a^\star$ stays hidden, and T3 is kept as context, not as a row on the image-only board.
On T1 the prompt contains $X$ and the instruction to fill the form, with repair notes, staff labels, and detector boxes absent.
On T2 the same prompt is joined by the highest non-view detections from an external one-stage detector~\cite{redmon2016yolo} and a short OCR line from the screen, and the prompt states that those boxes are predictions.
On T3 a declared agent may read seller repair text.
The oracle is scored after it has been shown $a^\star$ and the detector boxes, so its number bounds what the form contains once those fields are visible.

\paragraph{The form score.}
Active items cover view coverage, the presence and severity of defects by region (display, body and frame, ports, camera), and staff fields (model, storage, battery, lock and biometrics, blockers).
If an item has no evidence on that order, its gate $g_j$ is off and the item leaves both the numerator and the denominator of the ratio.
With awarded points $\hat{s}_j\in[0,M_j]$,
\begin{equation}
\mathrm{Score100}=100\cdot\frac{\sum_j g_j\hat{s}_j}{\sum_j g_j M_j}.
\label{eq:score}
\end{equation}
The leaderboard averages Eq.~\eqref{eq:score} over $\mathcal{A}$, which is Eq.~\eqref{eq:official}.
Splitting the items into appearance indices $\mathcal{Y}$ and staff indices $\mathcal{S}$ yields the two fill rates
\begin{equation}
\mathrm{Y\text{-}fill}=\frac{\sum_{j\in\mathcal{Y}} g_j\hat{s}_j}{\sum_{j\in\mathcal{Y}} g_j M_j},\qquad
\mathrm{S\text{-}fill}=\frac{\sum_{j\in\mathcal{S}} g_j\hat{s}_j}{\sum_{j\in\mathcal{S}} g_j M_j},
\label{eq:fills}
\end{equation}
which use the same gates and maxima as Score100 and therefore are not a second rank, but a decomposition that identifies which half of Eq.~\eqref{eq:score} moved.

Box F1 is a separate matching score.
Reference boxes and predicted boxes are paired greedily by intersection over union (IoU), and a pair is kept only when the classes agree and the overlap is at least $1/2$.
With $n_{\mathrm{tp}}$ kept pairs, $n_{\mathrm{fp}}$ unmatched predictions, and $n_{\mathrm{fn}}$ unmatched references,
\begin{equation}
F1=\frac{2n_{\mathrm{tp}}}{2n_{\mathrm{tp}}+n_{\mathrm{fp}}+n_{\mathrm{fn}}}.
\label{eq:f1}
\end{equation}
A box that does not parse counts in $n_{\mathrm{fp}}$, and $F1$ is reported next to Score100 without being added into Eq.~\eqref{eq:score}, so a system can improve localization and still lose the form.
If the seller omitted a view, every item that needs that view has $g_j{=}0$ and leaves both sums.
If the screen is present and storage is left blank, that staff item stays active with $\hat{s}_j{=}0$ while $M_j$ remains in the denominator, so the ratio falls.
Coverage counts whether a form came back, whereas Score100, Y-fill, and S-fill say what that form got right on $\mathcal{A}$.
Box F1 can rise while Score100 does not, which is the pattern a detector sidecar is allowed to produce.

\paragraph{The test cut and the three-image pack.}
The reported test split has 335 orders, dominated by iPhone listings; other device families are few and are reported only as exploratory slices.
Order identifiers also carry a listing-condition token (mint, light wear, heavy wear, outer crack, inner crack) that covers 327 of those orders; the remaining eight use a leftover token and are left out of the token-level means.
At evaluation every system in one protocol receives the same pack of at most three images with long side 448: a deterministic priority over exterior slots (front, back, side/bottom), never using staff labels or model outputs to choose views, and dropping a slot only if that view is missing.
A robustness slice of 80 orders is re-scored under blur, a missing view, an unreadable screenshot, and a modality conflict, and a device slice scores the 33 iPhone~11 orders inside the 335.
Neither slice replaces the 335-order leaderboard.
Training uses other orders, and an audit on order identifiers and on 256-bit secure hashes (SHA-256) of the images finds no overlap with the 335.
Every number reported below is computed on this cut.
Section~\ref{sec:lora} then studies a field-only adapter as a case study on the same board.

\section{Field-only LoRA}
\label{sec:lora}

\begin{figure}[t]
\centering
\includegraphics[width=\linewidth]{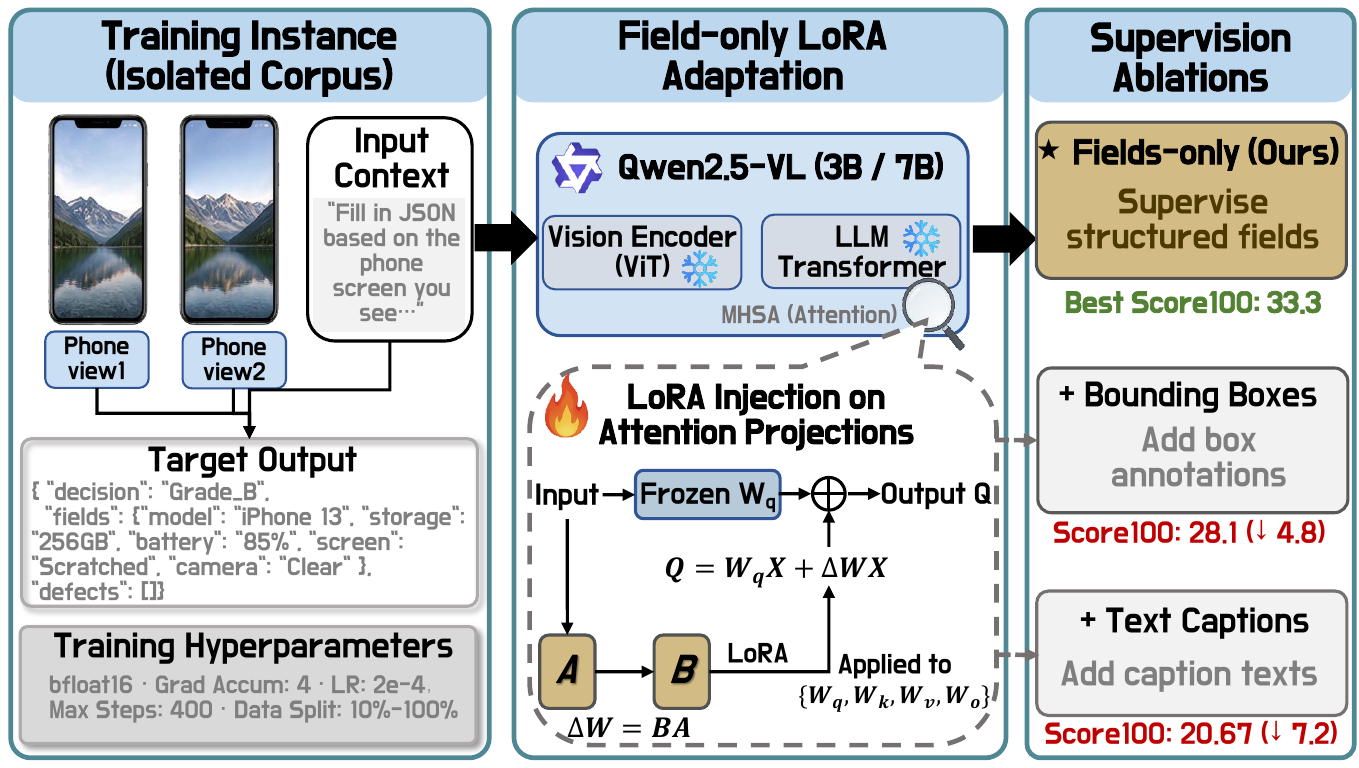}
\caption{Field-only LoRA on frozen Qwen2.5-VL.
Left: an isolated-corpus instance asks for an inspection JSON. The selected target fills the fields and leaves defects empty.
Center: LoRA on $W_q,W_k,W_v,W_o$ with $r{=}8$ and $\alpha{=}16$. The vision encoder and the transformer stay frozen.
Right: the box target and the caption target are ablations.
Measured scores are in Table~\ref{tab:sup}. The figure marks fields-only as the recipe we keep.}
\label{fig:adapt}
\end{figure}

The benchmark asks for the staff half of the form from the same photographs a zero-shot model is allowed to see.
The supervised span must write fields that Score100 pays for, namely model, storage, battery, lock and biometrics, and blockers when gated, while the test-time input stays identical to T1 with no detector sidecar and no staff labels.
That choice is deliberate. Appearance items already have strong zero-shot priors on this cut, whereas the staff half is where answered Score100 moves once an adapter is trained.
The case study therefore treats field writing as the trainable signal and keeps localization off the loss on the same T1 photographs.
Figure~\ref{fig:adapt} shows the supervised record, the frozen backbone with low-rank updates, and the two heavier targets kept only as ablations~\cite{hu2022lora,liu2023llava}.

\paragraph{Problem formulation.}
Let $f_\theta$ denote a vision-language model with parameters $\theta$, and write the T1 prompt as $(X,I)$.
Score100, Y-fill, S-fill, and box F1 are discontinuous in the discrete tokens of $f_\theta$, so we do not differentiate Score100 directly.
We seek a small update that raises the answered form score without changing $X$, freeze $\theta$ outside a low-rank residual, and supervise only the staff fields, the half of Eq.~\eqref{eq:score} that zero-shot models leave most incomplete on this cut.
The selected target $y^\star$ serializes those fields and closes the defect list as empty:
\begin{equation}
\min_{\Delta}\;\mathbb{E}_{(X,y^\star)}\Bigl[-\sum_{t=1}^{T}\log p_{\theta+\Delta}\bigl(y^\star_t\mid y^\star_{<t},X,I\bigr)\Bigr],
\label{eq:obj}
\end{equation}
where $\Delta$ is constrained to the LoRA factors below.
Eq.~\eqref{eq:obj} is a token-level surrogate for Score100. Staff-field tokens can enter S-fill, while box coordinates that the rank does not add are left out of $y^\star$.
Because the form score is gated and averaged only on answered orders, the objective must improve the filled fields without collapsing coverage.
A successful run is therefore one that raises answered Score100 and S-fill together, at coverage high enough to remain comparable to the zero-shot row. It is not a run that only raises box F1.
Ablations reintroduce boxes and captions under the same $\Delta$ budget so that a drop can be blamed on the supervised text and not on a weaker adapter.
This keeps the comparison inside one training budget and one image contract.

\paragraph{Frozen backbone and LoRA adapters.}
The backbone is Qwen2.5-VL-Instruct at 3B and 7B~\cite{qwen25vl2025}.
We load it in bfloat16, freeze the vision encoder, and attach LoRA only to $W_q,W_k,W_v,W_o$ with dropout $0.05$ and no bias~\cite{hu2022lora}.
Freezing the encoder keeps the visual features shared with the zero-shot baseline, so a Score100 gain can be attributed to how the language stack attends those features when writing staff keys, not to a new vision tower.
On a frozen projection $W$ the adapted map is
\begin{equation}
W'x = Wx + \frac{\alpha}{r}BAx,
\label{eq:lora}
\end{equation}
with $A\in\mathbb{R}^{r\times d}$, $B\in\mathbb{R}^{d\times r}$, $r{=}8$, and $\alpha{=}16$.
$B$ starts at zero, so the first forward pass matches the frozen checkpoint. Every other weight receives no gradient.
The rank and scale follow a light adapter setting that fits the short staff JSON without expanding the trainable footprint into the vision stack.
Only attention projections are updated, which limits drift in the shared visual stream while still allowing the language head to rewrite staff keys.
Any later box-F1 rise on the empty-defect recipe is produced by $W'$, not by the sidecar $D$.

\paragraph{Supervised target and empty defects.}
The assistant JSON has keys decision, fields, and defects.
Decision is answer when a usable staff form exists. Fields carries the gated staff half of $a^\star$. Defects is empty, so no coordinate tokens enter the loss.
The box ablation fills reference boxes. The caption ablation appends sentences never parsed into Score100.
Both keep the same $W'$, step budget, and T1 images.
Under Eq.~\eqref{eq:loss}, extra box or caption tokens dilute the staff-field signal that S-fill depends on, which is why the empty list concentrates capacity on the half that moves on this cut.
Leaving defects empty also matches the image-only contract at test time, where no detector boxes are available to copy.
The empty list is not a claim that localization is useless. It is a claim about what the official rank can reward under a fixed step budget.

\paragraph{Fine-tuning and T1 inference.}
Each instance packs two or three images with long side 448, a short instruction, and $y^\star$.
The chat template forms one causal sequence. The practical loss is
\begin{equation}
\mathcal{L}=-\frac{1}{T}\sum_{t=1}^{T}\log p(y_t\mid y_{<t}, X),
\label{eq:loss}
\end{equation}
with padding ignored and the frozen encoder blocked, so updates enter only through $A$ and $B$.
We train with Hugging Face Trainer, batch size 1, four accumulation steps, learning rate $2\times10^{-4}$, bfloat16, and gradient checkpointing, for 400 steps on the 10\% slice for about one epoch, and the same step budget at 25\%, 50\%, and 100\%.
Each fraction is one logged run under that fixed budget. It is not an epoch-matched or multi-seed estimate. Holding steps fixed while growing the slice is how coverage collapse under this budget becomes visible.
Training orders are isolated from the 335-cut by order id and image hash. Only adapters are saved.
No training image is reused in the 335-order board, so a lift on that board cannot be explained by memorizing the test photographs.
At test time adapters merge into $W_q,W_k,W_v,W_o$, decoding is greedy under the exact T1 prompt with no $D$ and no $a^\star$, and the parser scores Score100, fills, box F1, and coverage under the same decision rule as every other system.
If the model abstains or requests human review, the order leaves the answered set $\mathcal{A}$ exactly as for a zero-shot or closed system.
The same parser and decision rule keep the adapted row on the same board as the open and closed baselines.
Reported lifts are therefore changes on that shared board, not private scores from a different prompt or a different image pack.

\section{Experiments}
\label{sec:exp}

\begin{figure*}[t!]
\centering
\includegraphics[width=\textwidth]{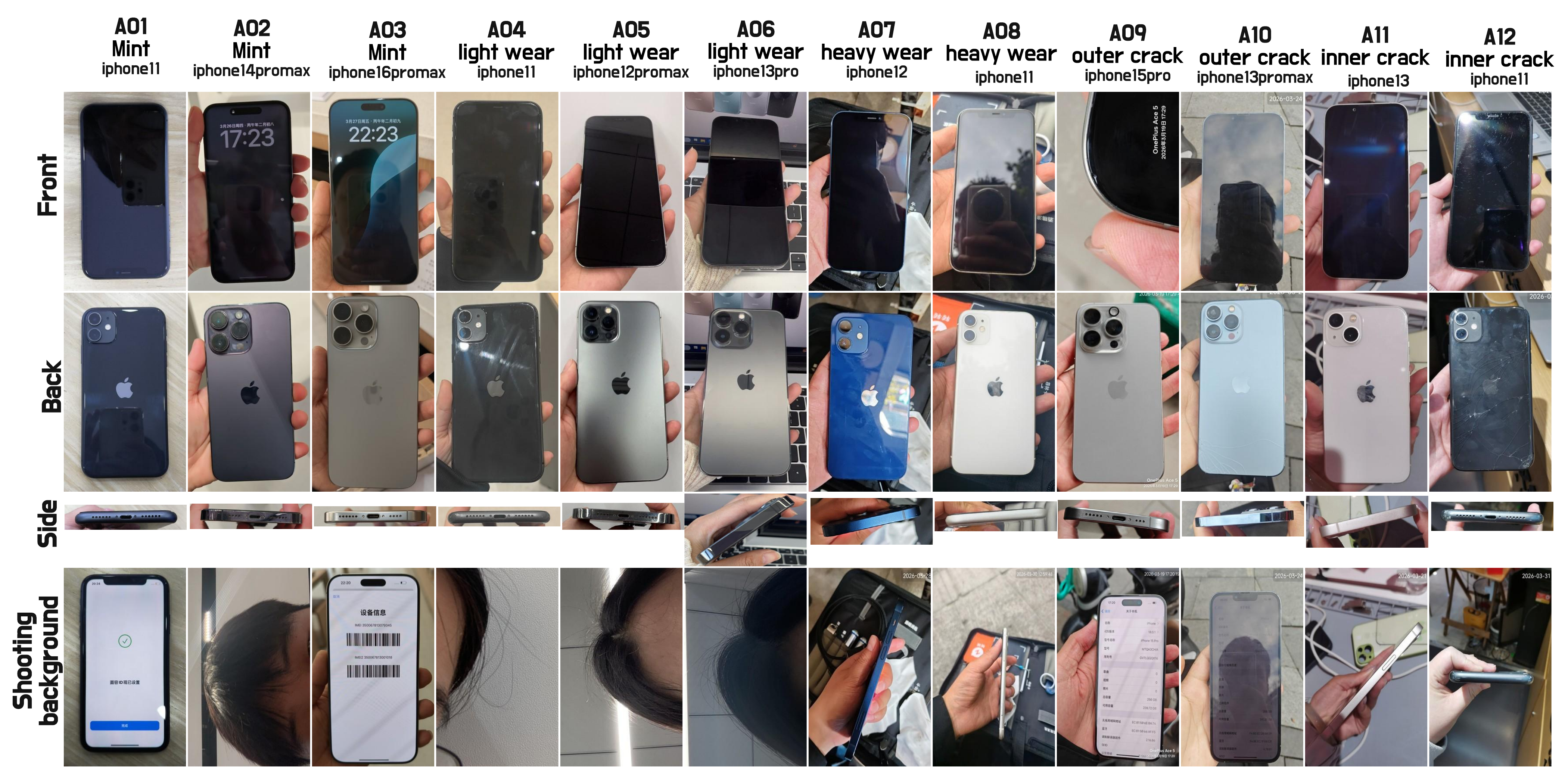}
\caption{Typical Order Cases gallery A01--A12.
Columns share a listing token and device family. Rows are front, back, side or bottom, and shooting-background exteriors.
Privacy-checked crops only.
Scores are in Table~\ref{tab:ap1}.}
\label{fig:ap1}
\end{figure*}

\begin{table}[t!]
\centering\scriptsize
\setlength{\tabcolsep}{5.5pt}
\renewcommand{\arraystretch}{0.96}
\caption{T1 scores for Figure~\ref{fig:ap1}.
zs, LoRA, and Step columns report Score100 with the decision ans, abs, or rev.
Device names are on the figure headers.
Row tint follows the listing token.}
\label{tab:ap1}
\begin{tabular}{@{}llccc l@{}}
\toprule
Case & Token & zs & LoRA & Step & Claim \\
\midrule
\rowcolor{qcSand}
\multicolumn{6}{@{}l@{}}{(A) Mint} \\
\rowcolor{qcSand}
A01 & mint & 15 (ans) & 25 (ans) & 25 (ans) & mint mid/low \\
\rowcolor{qcSand}
A02 & mint & 29 (ans) & 34 (ans) & 34 (ans) & LoRA +5 vs zs \\
\rowcolor{qcSand}
A03 & mint & 29 (ans) & 34 (ans) & 34 (ans) & LoRA +5 vs zs \\
\midrule
\rowcolor{qcTeal}
\multicolumn{6}{@{}l@{}}{(B) Light wear} \\
\rowcolor{qcTeal}
A04 & light & 19 (ans) & 25 (ans) & 25 (ans) & LoRA +5 vs zs \\
\rowcolor{qcTeal}
A05 & light & 19 (ans) & 25 (ans) & 19 (ans) & LoRA +5 vs zs \\
\rowcolor{qcTealBest}
A06 & light & 12 (ans) & 32 (ans) & 29 (rev) & Step to review \\
\midrule
\rowcolor{qcSlate}
\multicolumn{6}{@{}l@{}}{(C) Heavy wear} \\
\rowcolor{qcSlate}
A07 & heavy & 19 (ans) & 27 (ans) & 34 (ans) & LoRA +8 vs zs \\
\rowcolor{qcSlate}
A08 & heavy & 29 (ans) & 34 (ans) & 34 (ans) & LoRA +5 vs zs \\
\midrule
\rowcolor{qcClay}
\multicolumn{6}{@{}l@{}}{(D) Outer / inner crack} \\
\rowcolor{qcClay}
A09 & outer & 29 (ans) & 34 (ans) & 29 (rev) & Step to review \\
\rowcolor{qcClay}
A10 & outer & 18 (ans) & 5 (abs) & 12 (ans) & LoRA abs, zs ans \\
\rowcolor{qcClay}
A11 & inner & 19 (ans) & 25 (ans) & 25 (ans) & hard bin \\
\rowcolor{qcClay}
A12 & inner & 12 (ans) & 5 (abs) & 12 (ans) & LoRA abs, zs ans \\
\bottomrule
\end{tabular}
\end{table}

\subsection{Experimental setup}
T1 and T2 share one output schema: a JSON form, a defect list, and a decision of answer, abstain, or human review.
Each order uses the same deterministic three-image pack with long side 448 for every system in a protocol.
Open models are Qwen2.5-VL-3B and 7B~\cite{qwen25vl2025} and InternVL2.5-4B~\cite{internvl25}. Closed T1 and T2 models are Qwen3-VL-Plus~\cite{qwen3vl2025}, GLM-4.5V~\cite{glm45v2025}, Step-1o-turbo-vision, Doubao-Seed-2.1-Pro, and HY-Vision-2.0-Instruct.
Where an API returns empty unless chain-of-thought is disabled, we disable it.
Field-only adapters in Section~\ref{sec:lora} are evaluated under T1 only. T2 is for frozen backbones with the detector sidecar.
All T1 and T2 scores below come from one logged decoding pass per system under the shared pack and parser.
Figure~\ref{fig:ap1} and Table~\ref{tab:ap1} show twelve privacy-checked orders with T1 scores for zs, field-only LoRA, and Step under decisions ans, abs, or rev.
Scores match a 335-cut order by listing token and device family and are not a second leaderboard.
Mint A01 stays mid to low from 15 to 25. A06 goes to human review under Step. A10 and A12 show LoRA abstaining while zs answers.
A second gallery C01 through C12 is in the supplement.

\subsection{Main comparison on the image-only protocol}
Image-only systems score between 26.61 and 33.33 on the 335-order cut in Table~\ref{tab:main}.
InternVL2.5-4B is the best open zero-shot model at 29.31 with coverage 0.997. Closed models score 30.44 to 31.65 yet answer only 0.588 to 0.824 of the orders.
Coverage must sit next to Score100. A closed model can look strong on the orders it keeps while sending the rest to human review.
Reading Step-1o at 31.65 beside the 3B adapter at 33.33 without the answered counts would compare a mean on 259 orders with a mean on 311.
The privileged oracle that sees staff fields and detector boxes scores 86.66 with staff fill 0.590. A move from 26.61 to 33.33 remains a gain on an open task, and the remaining gap shows that T1 supervision is not enough to replace the quote or to close the staff half of the form.
At 10\% of the isolated corpus, the 3B field-only adapter reaches 33.33 at coverage 0.928 and the 7B adapter reaches 32.74 at coverage 1.000. The same 3B recipe at 25\% scores 29.27 at coverage 0.985.
The 10\% 3B row is the strongest high-coverage open run among the schedules logged here.
Appearance fill stays between 0.229 and 0.270. S-fill moves from 0.313 to 0.479, but S10 moves from 0.122 to 1.000 because every reference blocker list is empty and an empty prediction scores full credit.
We therefore treat composite S-fill with care and point to Score100 without S10, plus model and storage items, in the supplement. Without S10 the 3B adapter still leads the open rows on this cut at 29.47 against 28.09 and 27.44.
On the 33 iPhone~11 orders the four open systems fall to 22.54, 23.38, 27.88, and 27.11, yet each adapter stays above its zero-shot counterpart.

\begin{table}[ht]
\centering\scriptsize
\setlength{\tabcolsep}{3.5pt}
\renewcommand{\arraystretch}{0.96}
\caption{T1 on 335 orders.
Score100 averages answered orders, and Cov. is coverage, the logged fraction of 335.
Y-fill is the appearance half, S-fill is the staff half, and S10 is the blocker item.
The best Score100 inside each learned block is in bold.
The 10\% 3B row is the strongest high-coverage open run among schedules logged here.
S10 is the empty-blocker item. See the supplement for Score100 without S10.
The oracle is not a T1 system.}
\label{tab:main}
\begin{tabular}{@{}lcccccc@{}}
\toprule
System & Score100 & Cov. & Y-fill & S-fill & Box F1 & S10 \\
\midrule
\rowcolor{qcSand}
\multicolumn{7}{@{}l@{}}{(A) Open zero-shot} \\
\rowcolor{qcSand}
Qwen2.5-VL-3B & 26.61 & 0.976 & 0.246 & 0.313 & 0.024 & 0.122 \\
\rowcolor{qcSand}
Qwen2.5-VL-7B & 27.90 & 0.946 & 0.229 & 0.396 & 0.022 & 0.558 \\
\rowcolor{qcSandBest}
InternVL2.5-4B & \textbf{29.31} & 0.997 & 0.257 & 0.377 & 0.099 & 0.503 \\
\midrule
\rowcolor{qcTeal}
\multicolumn{7}{@{}l@{}}{(B) Field-only LoRA, image-only test} \\
\rowcolor{qcTealBest}
3B fields 10\% & \textbf{33.33} & 0.928 & 0.270 & 0.479 & 0.113 & 1.000 \\
\rowcolor{qcTeal}
3B fields 25\% & 29.27 & 0.985 & 0.251 & 0.390 & 0.097 & 0.994 \\
\rowcolor{qcTeal}
7B fields 10\% & 32.74 & 1.000 & 0.263 & 0.478 & 0.104 & 0.991 \\
\midrule
\rowcolor{qcSlate}
\multicolumn{7}{@{}l@{}}{(C) Closed models, on the orders they answer} \\
\rowcolor{qcSlate}
Qwen3-VL-Plus & 31.25 & 0.588 & 0.252 & 0.452 & 0.005 & 0.873 \\
\rowcolor{qcSlate}
GLM-4.5V & 30.44 & 0.824 & 0.249 & 0.433 & 0.025 & 0.855 \\
\rowcolor{qcSlateBest}
Step-1o-turbo-vision & \textbf{31.65} & 0.773 & 0.254 & 0.462 & 0.050 & 0.915 \\
\rowcolor{qcSlate}
Doubao-Seed-2.1-Pro & 30.66 & 0.699 & 0.237 & 0.467 & 0.047 & 0.889 \\
\rowcolor{qcSlate}
HY-Vision-2.0 & 30.64 & 0.618 & 0.238 & 0.465 & 0.019 & 0.966 \\
\midrule
\rowcolor{qcGold}
\multicolumn{7}{@{}l@{}}{(D) Ceiling, not a T1 system} \\
\rowcolor{qcGold}
Oracle (staff + det.) & 86.66 & 1.000 & 0.986 & 0.590 & 1.000 & 1.000 \\
\bottomrule
\end{tabular}
\end{table}

\begin{figure}[ht]
\centering
\includegraphics[width=\linewidth]{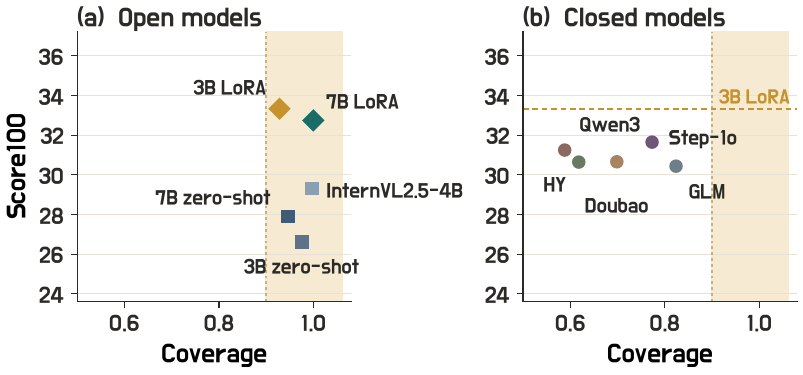}
\caption{Score100 against coverage on T1.
(a) Open zero-shot models and the 10\% field-only adapters.
(b) Closed models, with the dashed line at the 3B adapter score.
The band is coverage of at least 0.9.
Only the adapters sit above 32 inside that band.
The 25\% adapter is omitted because the point would overlap InternVL2.5-4B (Table~\ref{tab:main}).}
\label{fig:coverage}
\end{figure}

Figure~\ref{fig:coverage} plots Score100 against coverage.
Open zero-shot models answer nearly every order and score 26.61 to 29.31. Closed models score 30.44 to 31.65 while answering 0.588 to 0.824. The 10\% field-only adapters are the only points above 32 with coverage above 0.9.
The 25\% run is left off the figure because it would overlap InternVL2.5-4B, while Table~\ref{tab:main} keeps the row.
Reporting both numbers is required. A board that kept only the answered score would hide the orders closed models skip.
On the 202-order subset answered by 3B zero-shot, InternVL, 3B LoRA, Step, and GLM together, LoRA remains ahead at 33.36 against Step 32.45. Without S10 the same order holds at 29.49 against 29.02.
Extended Image-Only Analyses in the supplement separate abstention from human review and print the full no-S10 tables.

\subsection{Score structure by listing condition and form items}
\begin{figure}[ht]
\centering
\includegraphics[width=\linewidth]{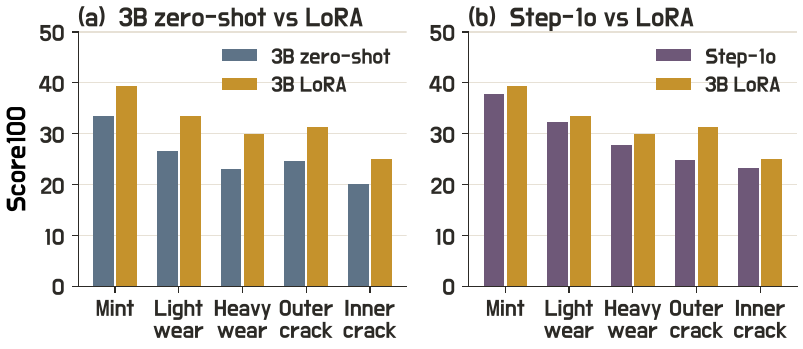}
\caption{Score100 by the listing-condition token on the order identifier.
(a) The 3B zero-shot model against field-only LoRA.
(b) Step-1o against the same adapter.
Both panels use the same scale and answered orders only.
The adapter is above the zero-shot model in every bin.
Answered counts are in the text.}
\label{fig:condition}
\end{figure}

Figure~\ref{fig:condition} averages Score100 inside the five listing tokens that cover 327 of the 335 orders.
For the 3B zero-shot model the answered counts are 70 mint, 124 light wear, 76 heavy wear, 31 outer crack, and 19 inner crack, with means 33.54, 26.57, 23.00, 24.66, and 20.04. Field-only LoRA is higher in every bin at 39.44, 33.48, 29.87, 31.41, and 25.11.
Step-1o falls from 37.90 to 23.29 on the orders it answers, and outer crack sits above heavy wear for the 3B zero-shot model at 24.66 against 23.00, so the tokens do not form a clean ladder.
Mint and inner crack still differ by 13.50 points, and the adapter is higher in both bins, including orders the listing already marks as damaged.
The pattern suggests that field-only LoRA improves the staff half inside each listing bin, but does not erase the visual difficulty of cracked devices. The mint-to-crack gap remains after adaptation, so the recipe is not a one-bin trick that buys mint by abandoning damaged uploads.

\begin{figure}[t]
\centering
\includegraphics[width=\linewidth]{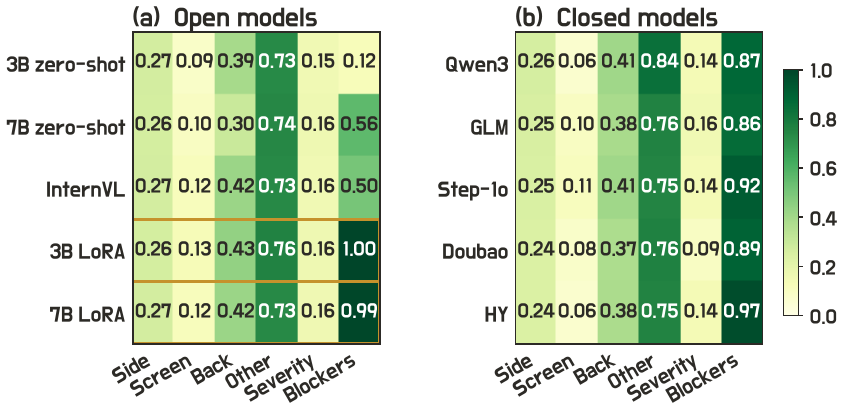}
\caption{Item F1 on answered orders.
(a) Open models. (b) Closed models.
Screen and severity stay pale.
The blocker cell separates the adapters from the 3B zero-shot model. The reference list is empty.
Gold outlines mark the adapters.}
\label{fig:items}
\end{figure}

\begin{figure}[t]
\centering
\includegraphics[width=\linewidth]{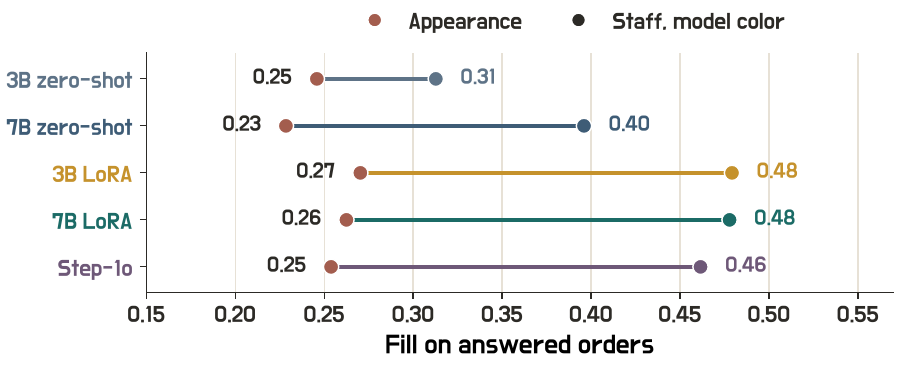}
\caption{The two references inside Score100, on one row per system.
The clay marker is appearance fill, and it stays near 0.25.
The colored marker is staff fill, and that is the end that rises for field-only LoRA.}
\label{fig:fills}
\end{figure}

\begin{figure}[t]
\centering
\includegraphics[width=0.95\linewidth]{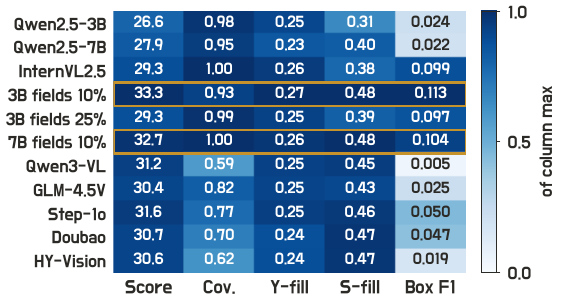}
\caption{T1 on one grid.
Each column is colored by its own maximum, the color bar is that share, and the printed number is the logged value.
Gold outlines mark the 10\% field-only adapters.
They are dark on score and staff fill together.
Box F1 does not become the dark column.}
\label{fig:board}
\end{figure}

Figure~\ref{fig:items} splits the form into items. Screen and severity stay low. The residual other region is already high before adaptation. The blocker cell moves from 0.122 to 1.000 because the reference list is empty and the adapter leaves it empty.
Implementation Notes in the supplement score a constant form that never sees the images. An empty form scores 21.19. A majority-field constant scores 28.13. The 3B adapter at 33.33 is 5.20 above that constant with staff fill 0.479 against 0.321.
That margin shows the adapted row is not a dressed-up majority predictor. Without looking at $X$, a constant cannot reach the staff fill that the field-only target trains.
Figure~\ref{fig:fills} shows appearance fill near 0.25 while staff fill rises from 0.313 to 0.479. That staff marker includes the empty-blocker credit discussed above.
Figure~\ref{fig:board} restores the 25\% adapter on the same T1 grid, with field-only LoRA at 10\% dark on score and coverage while box F1 stays light.
Screen and severity staying pale is consistent with the empty-defect target. The loss never asks for region boxes that would support those appearance cells.

\subsection{Supervision ablation}
Fields-only supervision raises Score100, while boxes and captions do not, as shown in Table~\ref{tab:sup} and Figure~\ref{fig:adapt}.
Fields-only at 10\% scores 33.33 at coverage 0.928.
Writing boxes lowers Score100 to 28.10 with box F1 0.105 against 0.113, and adding captions drops the score to 20.67 at coverage 0.296.
The box ablation does not buy a higher box F1 that compensates for the lost form score, and the caption ablation collapses coverage.
That outcome is the experimental counterpart of the dilution argument in Section~\ref{sec:lora}. When the supervised span spends tokens on coordinates or captions that Score100 ignores, the answered form score falls even though the adapter geometry is unchanged.
At 50\% the answered score is 31.00 with coverage 0.537. At 100\% coverage falls to 0.045 with score 13.70. The 25\% run reaches coverage 0.985 with score 29.27, so the 10\% lead is not a trick of skipping hard orders.
Stars mark coverage below 0.9.
Under the fixed 400-step budget, lengthening $y^\star$ or growing the corpus without more steps reduces how much of the loss lands on staff-field tokens.
We hold the step budget fixed and report single runs. The 10\% lead is the best among these logged configs. It is not a multi-seed training law.

\begin{table}[t]
\centering\scriptsize
\setlength{\tabcolsep}{3.5pt}
\renewcommand{\arraystretch}{0.96}
\caption{Qwen2.5-VL-3B on T1.
(A) Supervision at 10\% of the corpus and 400 steps.
(B) Fields-only as the corpus fraction grows.
The best answered score that still returns a form is in bold.
Stars mark runs whose coverage is too low to treat the score as a leaderboard entry.}
\label{tab:sup}
\begin{tabular}{@{}lccc@{}}
\toprule
Recipe & Score100 & Cov. & Box F1 \\
\midrule
\rowcolor{qcTeal}
\multicolumn{4}{@{}l@{}}{(A) Target at 10\%} \\
\rowcolor{qcTealBest}
Fields only (ours) & \textbf{33.33} & 0.928 & 0.113 \\
\rowcolor{qcSand}
Fields + boxes & 28.10 & 0.964 & 0.105 \\
\rowcolor{qcClay}
Fields + boxes + captions & 20.67 & 0.296$^\star$ & 0.071 \\
\midrule
\rowcolor{qcSlate}
\multicolumn{4}{@{}l@{}}{(B) Fields-only fraction} \\
\rowcolor{qcSlate}
Fields only, 25\% & 29.27 & 0.985 & 0.097 \\
\rowcolor{qcClay}
Fields only, 50\% & 31.00$^\star$ & 0.537 & 0.061 \\
\rowcolor{qcClay}
Fields only, 100\% & 13.70$^\star$ & 0.045 & 0.000 \\
\bottomrule
\end{tabular}
\end{table}

\subsection{Effect of the detector sidecar}
T2 asks whether predicted boxes and a short screen transcript help the form while $a^\star$ stays hidden.
Holding images, form instruction, and decoding fixed and appending only $D$ repeats the same split in Table~\ref{tab:t2}. Box F1 rises on every backbone. Score100 falls by 3.1, 6.8, and 2.9 on the open and Qwen3 rows and by 1.9 on Step-1o and HY-Vision. GLM-4.5V changes by $+0.14$ in a single logged run, which we treat as a descriptive bump and not as a claimed gain.
Doubao-Seed-2.1-Pro sends 62.7\% of T2 orders to human review at coverage 0.173, so its T2 score is omitted and is not ranked next to rows that answer almost the full cut.
On the 7B model, staff fill falls from 0.396 to 0.182 while box F1 rises from 0.022 to 0.103.
These are protocol-level observations under the logged sidecar. Because $D$ also lengthens the prompt, we do not attribute the drop to boxes alone.
A sidecar therefore cannot take the T1 rank in these logs, and it cannot replace the field-only update that never receives $D$ at test time.
The T1 board remains the place where image-only systems are compared.
Grouped bars and staff-fill columns are expanded in the supplement.

\begin{table}[t]
\centering\scriptsize
\setlength{\tabcolsep}{3.5pt}
\renewcommand{\arraystretch}{0.96}
\caption{The same backbone on T1 and T2 with images and decoding held fixed.
$\Delta$ is the change in Score100 from a single logged run.
Box F1 rises on every row. GLM-4.5V's $+0.14$ is descriptive only.
Doubao T2 coverage is 0.173, so its T2 score and the change are marked n.a.}
\label{tab:t2}
\begin{tabular}{@{}lccccc@{}}
\toprule
Backbone & T1 & T2 & $\Delta$ & T1 F1 & T2 F1 \\
\midrule
\rowcolor{qcSand}
\multicolumn{6}{@{}l@{}}{(A) Open zero-shot} \\
\rowcolor{qcClay}
Qwen2.5-VL-3B & 26.61 & 23.50 & $-$3.1 & 0.024 & 0.105 \\
\rowcolor{qcClay}
Qwen2.5-VL-7B & 27.90 & 21.13 & $-$6.8 & 0.022 & 0.103 \\
\midrule
\rowcolor{qcSlate}
\multicolumn{6}{@{}l@{}}{(B) Closed models} \\
\rowcolor{qcClay}
Qwen3-VL-Plus & 31.25 & 28.36 & $-$2.9 & 0.005 & 0.081 \\
\rowcolor{qcClay}
GLM-4.5V & 30.44 & 30.58 & $+0.14$ & 0.025 & 0.116 \\
\rowcolor{qcClay}
Step-1o & 31.65 & 29.72 & $-$1.9 & 0.050 & 0.082 \\
\rowcolor{qcSlate}
Doubao-Seed-2.1 & 30.66 & n.a. & n.a. & 0.047 & 0.089 \\
\rowcolor{qcClay}
HY-Vision-2.0 & 30.64 & 28.70 & $-$1.9 & 0.019 & 0.067 \\
\bottomrule
\end{tabular}
\end{table}

\subsection{Robustness to visual corruptions}
Removing images on a separate 100-order check leaves 3B zero-shot coverage at 0.05, while an image-only run scores 26.54 next to 25.73 with the full prompt, so the images carry the form.
Table~\ref{tab:robust} then rescores one fixed 80-order stress slice under four corruptions applied to the same identifiers with $n{=}80$ each. Overall, zero-shot falls to coverage 0.55 and Score100 24.04, while the 10\% field-only model keeps coverage 0.86 and scores 26.13.
Blur is the largest coverage gap at 0.35 against 1.00. Missing view raises the LoRA score from 24.66 to 31.25. An unreadable shot still favors the adapter on both score and coverage.
On modality conflict the LoRA score is lower at 22.69 against 24.76 even though coverage rises at 0.80 against 0.65.
That cell is descriptive only. When streams disagree, answering more often does not guarantee a higher answered mean.
The overall row still favors the adapted system on this slice. Corruptions can change visible evidence and the decision to answer, so the table is a stress reading and not a multi-seed interval.
Decision-rate bars for the same slice are in the supplement.

\begin{table}[t]
\centering\scriptsize
\setlength{\tabcolsep}{3.5pt}
\renewcommand{\arraystretch}{0.96}
\caption{Robustness on 80 orders.
Score100 is the mean on answered orders.
The LoRA columns are the 3B field-only model trained on 10\% of the corpus.
Bold marks the higher score inside each condition.}
\label{tab:robust}
\begin{tabular}{@{}lcccc@{}}
\toprule
 & \multicolumn{2}{c}{3B zero-shot} & \multicolumn{2}{c}{3B LoRA 10\%} \\
\cmidrule(lr){2-3}\cmidrule(lr){4-5}
Condition & Score & Cov. & Score & Cov. \\
\midrule
\rowcolor{qcTeal}
Overall & 24.04 & 0.55 & \textbf{26.13} & 0.86 \\
\rowcolor{qcTeal}
Blur & 23.29 & 0.35 & \textbf{25.89} & 1.00 \\
\rowcolor{qcTeal}
Missing view & 24.66 & 0.70 & \textbf{31.25} & 0.80 \\
\rowcolor{qcTeal}
Unreadable shot & 22.74 & 0.50 & \textbf{24.82} & 0.85 \\
\rowcolor{qcClay}
Modality conflict & \textbf{24.76} & 0.65 & 22.69 & 0.80 \\
\bottomrule
\end{tabular}
\end{table}

\section{Conclusion}
We present PhoneQC as a 335-order form-score benchmark for multimodal trade-in inspection under matched protocols, with field-only LoRA as a case study on the same T1 photographs.
The board keeps Score100 with coverage on this iPhone-heavy cut.
The 3B adapter reaches 33.33 at coverage 0.928 among the tested schedules; without the empty-blocker item it remains ahead of the open baselines on this cut, while boxes, captions, and the logged sidecar leave the oracle gap at 86.66 open.
We hope this encourages further order-level inspection benchmarks beyond the present split.

\section*{LLM Usage Statement}
During the writing of this paper, we used ChatGPT to assist with formatting and language polishing. The authors are responsible for all content.

{
    \small
    \bibliographystyle{unsrtnat}
    \bibliography{main}
}

\clearpage
\appendix

\begin{center}
{\Large\bfseries Supplementary Material}
\end{center}
\vspace{0.75em}

\section{Overview}
This supplement expands the main-paper analyses on the same 335-order cut.
Score100 is always reported with coverage, and a star marks runs with coverage below 0.9.
Notation follows the main paper: T1 is image-only, T2 adds a detector sidecar, LoRA is low-rank adaptation, and box F1 is not added into Score100.
The sections below cover broader impact and assets, typical order cases, extended image-only analyses, the detector sidecar and robustness, adaptation ablations and slices, the appraiser agreement study, sensitivity without S10 and common-answered subsets, implementation and reproducibility notes, and scope and reporting.
The release package includes CSV tables, scored rows for the 50 agreement orders, and privacy-checked exterior views where a safe crop exists, but not the full corpus or personal identifiers.

\section{Broader Impact and Assets}
\paragraph{Broader impact.}
PhoneQC scores whether a system filled a staff form from the images an upload already contains.
A higher Score100 is neither a repurchase price nor a quote a desk can send.
The best image-only score in these logs is 33.33, and the staff-field oracle is 86.66: the distance between those two numbers is the room in which a person still writes the quote.
Using the answered mean alone can hide that fact, because a model may look strong on the orders it is willing to answer and send the rest to review or return nothing.
The coverage column is therefore part of the result.

The photographs can show hands, a work surface, and nearby objects.
Ethics consent forms for the orders were obtained.
This package does not release the corpus.
It releases privacy-checked exterior crops in the main-paper A01--A12 gallery and in the supplement C01--C12 gallery (Figure~\ref{fig:ap2}), with screen captures, serial numbers, account UI, and price slips withheld, so that the kind of photograph scored by the contract remains visible without exposing customer text.

\paragraph{Released assets.}
The open backbones are used under the licenses published with their checkpoints.
Qwen2.5-VL-3B and Qwen2.5-VL-7B are covered by the Qwen Research License shipped with those weights, and InternVL2.5-4B is used under the license shipped with that checkpoint.
Neither set of weights is included here.
The closed systems are API endpoints whose weights are not ours to release, and the package records only the scores those endpoints returned under the T1 and T2 contracts.

Apart from the main-paper A01--A12 gallery, the supplement C01--C12 gallery (Figure~\ref{fig:ap2}), and the three sample views in Figure~\ref{fig:views}, the remaining order photographs and the staff forms are not in this package.
What is included are the aggregate tables printed below as CSV, one row per order with a SHA-256 hash of the order identifier, a 50-row case table with staff tokens and logged T1 scores for the Appraiser Agreement Study, and the per-column Score100 table under Figure~\ref{fig:ap2}.
The hash is the isolation key used by the split, not a customer record.

\paragraph{Relation to existing datasets.}
The cut contains 335 trade-in orders because the evaluation unit is an order, not a single photograph: each order has several views, a staff form, and a rule about which fields a system may see.
A screen-glass set such as SSGD is one view of one pane~\cite{ssgd2023}.
MVTec AD~\cite{mvtec2019}, VisA~\cite{visa2022}, and Kaputt~\cite{kaputt2025} are anomaly sets under controlled capture and do not ask for a staff form.
VQA~\cite{antol2015vqa}, DocVQA~\cite{docvqa2021}, ScreenQA~\cite{screenqa2025}, MMBench~\cite{mmbench2024}, MMMU~\cite{mmmu2024}, and SEED-Bench~\cite{li2024seed} score a question, a document, a screen, or an exam item.
None of those resources fixes a forbidden-field contract for a multi-view trade-in order, so we did not run the benchmark on them: a number on a different task would not answer the question this cut is built for.

The reference on this cut is the staff form, not a score from a vision-language model.
An automatic labeler would have raised a second question, whether the labeler agrees with a person, before any system could be ranked.
The Appraiser Agreement Study below is that check on the staff form itself: twenty-four appraisers filled the same 50 orders, the rank uses the staff record, and the study reports where that record is stable and where it is soft.

\setcounter{figure}{8}

\section{Typical Order Cases}
\label{sec:cases}
The main paper prints gallery A01--A12 with its score table.
This section keeps the second pack C01--C12 so the listing axis is not tied to one device set.
Notation matches the main paper: zs is Qwen2.5-VL-3B zero-shot, LoRA is field-only LoRA at 10\%, Step is Step-1o-turbo-vision, and ans / abs / rev are answer, abstain, and human review.
Scores are logged T1 values for a 335-cut order that shares the column listing token and the closest device family in the header; identifiers stay hashed in the release CSV.

\begin{figure*}[t]
\centering
\includegraphics[width=\textwidth]{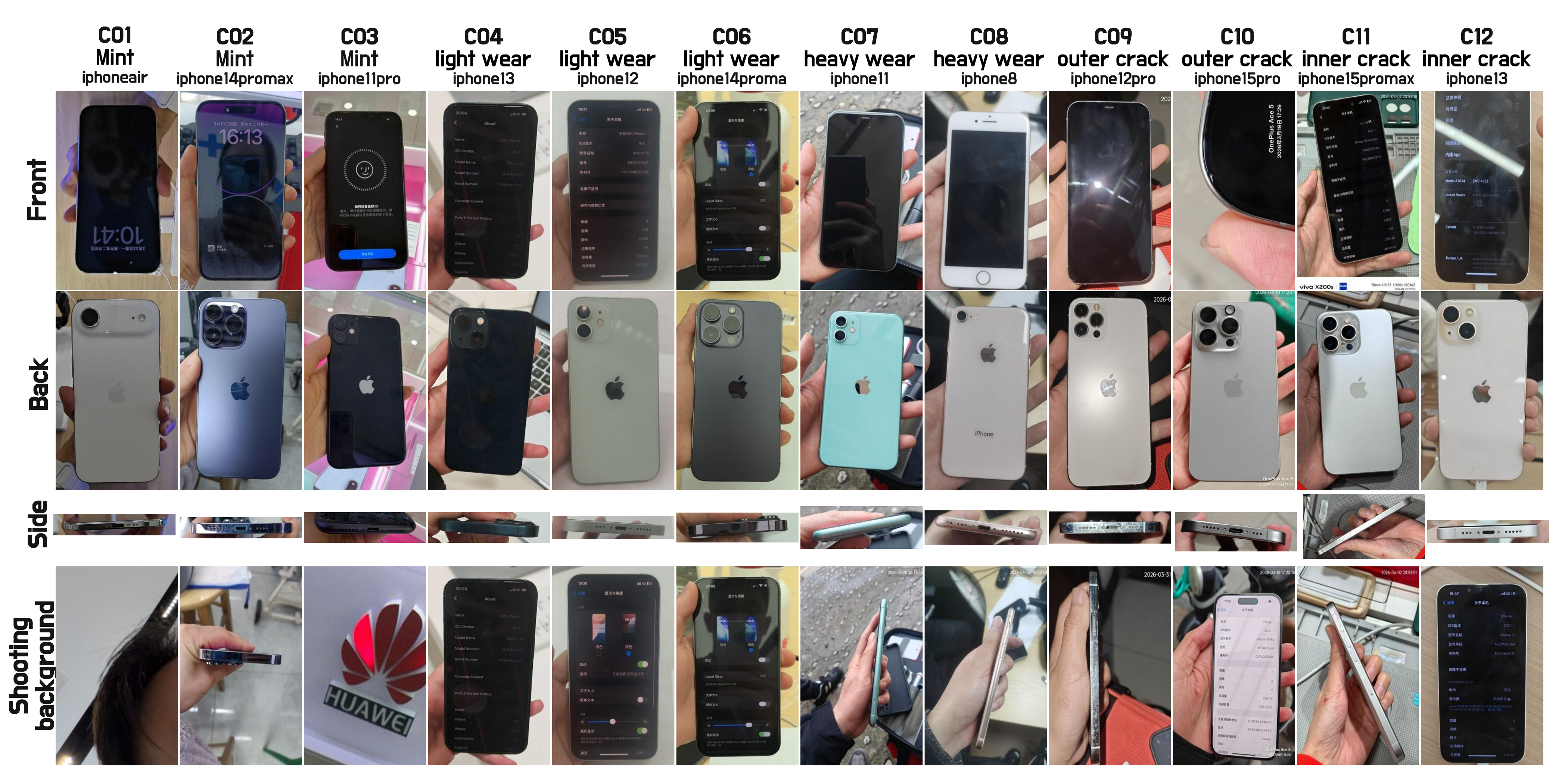}
\caption{Typical Order Cases gallery C01--C12.
Same $4\times12$ layout as the main-paper A01--A12 gallery, with a second set of devices under the five listing tokens.
Privacy-checked crops only.}
\label{fig:ap2}
\end{figure*}

\begin{table*}[t]
\centering\footnotesize
\setlength{\tabcolsep}{3.5pt}
\caption{T1 scores for the twelve columns of Figure~\ref{fig:ap2}.
zs / LoRA / Step are Score100 with decision ans, abs, or rev.}
\label{tab:ap2}
\begin{tabular}{@{}llcclll@{}}
\toprule
Case & Token & Device & zs & LoRA & Step & Claim \\
\midrule
C01 & mint & iphoneair & 29 (ans) & 34 (ans) & 30 (ans) & LoRA +5 vs zs \\
C02 & mint & iphone14promax & 29 (ans) & 34 (ans) & 34 (ans) & LoRA +5 vs zs \\
C03 & mint & iphone11pro & 32 (ans) & 37 (ans) & 37 (ans) & LoRA +5 vs zs \\
C04 & light wear & iphone13 & 19 (ans) & 25 (ans) & 25 (ans) & LoRA +5 vs zs \\
C05 & light wear & iphone12 & 15 (ans) & 25 (ans) & 25 (ans) & LoRA +10 vs zs \\
C06 & light wear & iphone14promax & 19 (ans) & 12 (ans) & 27 (ans) & Step +8 vs zs \\
C07 & heavy wear & iphone11 & 29 (ans) & 34 (ans) & 34 (ans) & LoRA +5 vs zs \\
C08 & heavy wear & iphone8 & 19 (ans) & 25 (ans) & 25 (ans) & LoRA +5 vs zs \\
C09 & outer crack & iphone12pro & 18 (ans) & 5 (abs) & 12 (ans) & LoRA abstains; zs answers \\
C10 & outer crack & iphone15pro & 29 (ans) & 34 (ans) & 29 (rev) & Step to human review \\
C11 & inner crack & iphone15promax & 29 (ans) & 34 (ans) & 34 (ans) & inner-crack hard bin \\
C12 & inner crack & iphone13 & 19 (ans) & 25 (ans) & 25 (ans) & inner-crack hard bin \\
\bottomrule
\end{tabular}
\end{table*}

\paragraph{Reading pack C01--C12.}
Mint C01--C03 show a modest LoRA lift rather than a solved form.
Light-wear C05 gains $+10$ Score100 with all three systems answering; C06 is the opposite shape, where Step leads zs while LoRA falls.
Outer-crack C09 reproduces the main-paper abstain pattern: LoRA abs at Score100 5 while zs still answers at 18.
C10 pairs an outer-crack header with Step human review.
Inner-crack C11--C12 stay mid/low even when LoRA is above zs, so the hard bin remains visible after adaptation.

\paragraph{Fifty-order CSV.}
The A01--A50 agreement pack remains in the release CSV with hashed order identifiers, staff tokens, and the same three T1 scores.
On that pack the mean Score100 is 26.49 for the 3B zero-shot model, 31.95 for field-only LoRA, and 32.00 for Step-1o; those means are a slice of the cut, not a second leaderboard.

\begin{figure}[t]
\centering
\includegraphics[width=0.32\linewidth]{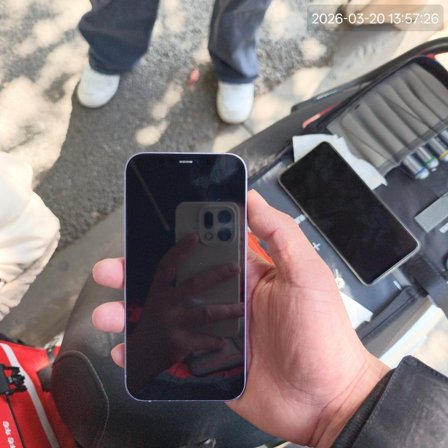}
\hfill
\includegraphics[width=0.32\linewidth]{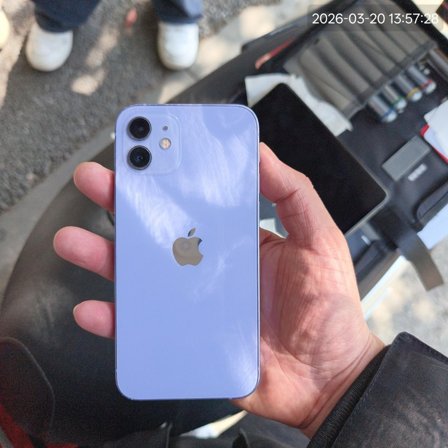}
\hfill
\includegraphics[width=0.32\linewidth]{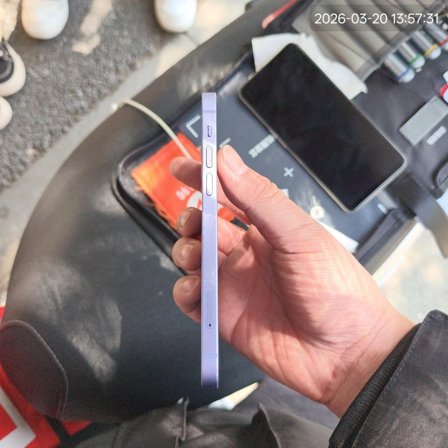}
\caption{Three exterior views of one order, long side 448.
Left to right: front, back, side.}
\label{fig:views}
\end{figure}

\paragraph{Sample views.}
\label{sec:views}
Figure~\ref{fig:views} shows one additional order after the long side is resized to 448 pixels, the same resize used in adapter training.
The three files are a front, a back, and a side held in a hand, with another phone and a work surface in the frame.
No screen capture is included.
Together with the main-paper A01--A12 gallery and Figure~\ref{fig:ap2}, these views illustrate the input contract; they are not a second test set.

\section{Extended Image-Only Analyses}
\begin{figure}[t]
\centering
\includegraphics[width=\linewidth]{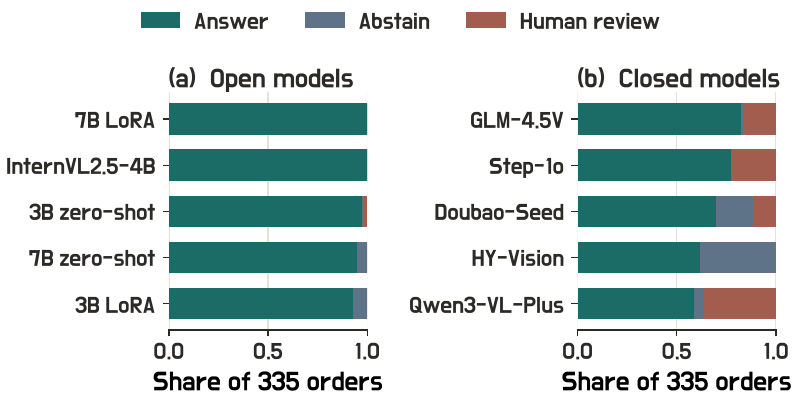}
\caption{Where the unanswered orders go.
(a) Open models. (b) Closed models.
Coverage in the main leaderboard is the teal answer share.
Closed models spend the rest on human review or abstention.}
\label{fig:s2}
\end{figure}

\paragraph{Decision breakdown.}
Coverage in the main leaderboard is a single number that adds abstention and human review together, yet those two decisions are not the same operational outcome: an abstention returns nothing a staff member can edit, while a human-review flag returns the order to a person.
Figure~\ref{fig:s2} separates them.
The teal share is the answer rate used as coverage in the rank, and the other two shares are the orders that never enter Score100.

Open models in panel (a) are almost entirely teal, while the closed models in panel (b) are not.
That contrast is why a closed Score100 in the low 30s cannot be placed next to an open score without coverage beside it.
The closed models score 30.44 to 31.65 while answering only 0.588 to 0.824 of the cut, whereas the 10\% field-only adapters sit above 32 and still answer 0.928 and 1.000.
If the board kept only the answered mean, the closed models would look close to the adapters and the orders they skip would disappear.

Table~\ref{tab:s-decision} is the count behind the bars. The three rates are shares of 335 and sum to 1. The answered column is the size of the set that Score100 actually averages.

\begin{table}[t]
\centering\footnotesize
\setlength{\tabcolsep}{3.5pt}
\caption{Answered counts behind Figure~\ref{fig:s2}.
The three rates are shares of 335 orders and sum to 1.}
\label{tab:s-decision}
\begin{tabular}{@{}lcccc@{}}
\toprule
System & Answered & Answer & Abstain & Review \\
\midrule
Qwen2.5-VL-3B & 327 & 0.976 & 0.003 & 0.021 \\
Qwen2.5-VL-7B & 317 & 0.946 & 0.054 & 0.000 \\
InternVL2.5-4B & 334 & 0.997 & 0.003 & 0.000 \\
3B LoRA fields 10\% & 311 & 0.928 & 0.072 & 0.000 \\
7B LoRA fields 10\% & 335 & 1.000 & 0.000 & 0.000 \\
Qwen3-VL-Plus & 197 & 0.588 & 0.051 & 0.361 \\
GLM-4.5V & 276 & 0.824 & 0.012 & 0.164 \\
Step-1o & 259 & 0.773 & 0.003 & 0.224 \\
Doubao-Seed & 234 & 0.699 & 0.185 & 0.116 \\
HY-Vision & 207 & 0.618 & 0.379 & 0.003 \\
\bottomrule
\end{tabular}
\end{table}

The open block fails rarely, and it fails in different small ways. The 3B zero-shot model answers 327 orders, so 8 orders never enter its mean, and almost all of that remainder is human review at 0.021. The 7B zero-shot model answers 317 and abstains on the other 18, with a review rate of 0. The 4B model leaves a single order out. Field-only LoRA on the 3B backbone answers 311 and abstains on 24, and it never sends an order to review. The 7B adapter answers all 335. The adapter's coverage cost, where it exists, is abstention, not a handoff. Its Score100 of 33.33 is therefore a mean on 311 orders, not a mean on a thin slice.

The closed block does not fail in one way, which is what panel (b) is for. Qwen3-VL-Plus answers 197 orders. The other 138 are mostly human review, at 0.361, with abstention only 0.051. Step-1o answers 259 and reviews 0.224, and it almost never abstains. GLM-4.5V answers 276, reviews 0.164, and abstains on 0.012. Doubao-Seed splits the remainder, with abstention 0.185 and review 0.116, and 234 orders in the mean. HY-Vision answers 207 and abstains on 0.379, the largest abstention share in the table, while its review rate is 0.003. A single coverage number would add those behaviors together. The figure keeps them apart, so a reader can see that one closed model is handing work to a person and another is returning no decision.

This split is what makes the comparison with the adapter fair.
Step-1o has the highest closed Score100, 31.65, on 259 orders, while the 3B adapter scores 33.33 on 311 orders and does not use review.
The adapter is ahead on the orders it answers and still returns a form on a larger share of the cut.
The closed mean is real on the orders inside its answered set, but it is not a score for the orders that set never included.

\paragraph{Extended T1 leaderboard.}
Table~\ref{tab:s-t1} repeats the T1 logs so the fill rates, the blocker item, and the 25\% adapter can be read in one place.
The 25\% row is the adapter omitted from the main-paper coverage scatter, because 29.27 at coverage 0.985 lands on InternVL2.5-4B at 29.31 and coverage 0.997.
The table keeps it: the fraction story is incomplete without that control.

\begin{table}[t]
\centering\scriptsize
\setlength{\tabcolsep}{3.5pt}
\caption{T1 on 335 orders, with the two fill terms and the blocker item.
Score100 averages answered orders.
Field-only LoRA is the row that is high on S-fill, S10, and coverage together, while box F1 stays near 0.1.
The 25\% row is the adapter omitted from the main-paper coverage scatter because it lands on InternVL2.5-4B.}
\label{tab:s-t1}
\begin{tabular}{@{}lcccccc@{}}
\toprule
System & Score100 & Cov. & Y-fill & S-fill & Box F1 & S10 \\
\midrule
Qwen2.5-VL-3B & 26.61 & 0.976 & 0.246 & 0.313 & 0.024 & 0.122 \\
Qwen2.5-VL-7B & 27.90 & 0.946 & 0.229 & 0.396 & 0.022 & 0.558 \\
InternVL2.5-4B & 29.31 & 0.997 & 0.257 & 0.377 & 0.099 & 0.503 \\
3B LoRA fields 10\% & 33.33 & 0.928 & 0.270 & 0.479 & 0.113 & 1.000 \\
3B LoRA fields 25\% & 29.27 & 0.985 & 0.251 & 0.390 & 0.097 & 0.994 \\
7B LoRA fields 10\% & 32.74 & 1.000 & 0.263 & 0.478 & 0.104 & 0.991 \\
\midrule
Qwen3-VL-Plus & 31.25 & 0.588 & 0.252 & 0.452 & 0.005 & 0.873 \\
GLM-4.5V & 30.44 & 0.824 & 0.249 & 0.433 & 0.025 & 0.855 \\
Step-1o-turbo-vision & 31.65 & 0.773 & 0.254 & 0.462 & 0.050 & 0.915 \\
Doubao-Seed-2.1-Pro & 30.66 & 0.699 & 0.237 & 0.467 & 0.047 & 0.889 \\
HY-Vision-2.0 & 30.64 & 0.618 & 0.238 & 0.465 & 0.019 & 0.966 \\
\bottomrule
\end{tabular}
\end{table}

Within the open zero-shot models, the 7B score is 27.90, 1.29 points above the 3B score of 26.61. Staff fill rises from 0.313 to 0.396, while appearance fill falls from 0.246 to 0.229. The S10 increase from 0.122 to 0.558 reflects the all-empty blocker reference. InternVL2.5-4B scores 29.31, with box F1 of 0.099 and coverage of 0.997; its staff fill is 0.377 and S10 is 0.503. Appearance fill remains near one quarter across these models.

The 3B adapter at 10\% scores 33.33 at coverage 0.928. Appearance fill is 0.270, within the 0.229--0.270 range of the learned rows. Staff fill is 0.479, up from 0.313, and S10 is 1.000, up from 0.122. Because S10 is constant in the references, this contributes to the staff-fill and Score100 increases. Box F1 is 0.113, close to InternVL's 0.099; the score difference is not explained by stronger localization.

The 25\% field-only run scores 29.27 at coverage 0.985, below the 10\% run despite answering more orders. Staff fill is 0.390 and appearance fill is 0.251; S10 remains 0.994. Under the fixed step budget, a larger training fraction is not automatically better. The 7B adapter at 10\% scores 32.74 at full coverage, with staff fill 0.478 and S10 0.991. The 3B 10\% run is the highest-scoring high-coverage configuration among those tested.

The closed rows sit in a narrow score band, 30.44 to 31.65, with appearance fill still 0.237 to 0.254. Their staff fill, 0.433 to 0.467, is near the adapter's 0.479, and their blocker items are already high, from 0.855 to 0.966. What they do not share with the adapter is coverage. The highest closed score, Step-1o at 31.65, is computed on coverage 0.773. The adapter's 33.33 is computed on coverage 0.928. Box F1 on these closed rows is 0.005 to 0.050, below the open adapter. A low box F1 with a mid-30s score is further evidence that the rank is not a localization contest.

The oracle has appearance fill 0.986 and staff fill 0.590. Its Score100 of 86.66 is a privileged upper bound, not 100. The 3B adapter remains far below this ceiling, with appearance errors still prominent.

\paragraph{Per-item decomposition.}
A single Score100 hides the item pattern drawn in the main-paper item heatmap.
Table~\ref{tab:s-items} prints those item F1 values: Y04 side, Y05 screen, Y06 back, Y07 residual other region, Y08 severity, and S10 blockers.

\begin{table}[t]
\centering\scriptsize
\setlength{\tabcolsep}{3.5pt}
\caption{Item F1, the numbers drawn in the main-paper item heatmap.
Y04 side, Y05 screen, Y06 back, Y07 other, Y08 severity, S10 blockers.}
\label{tab:s-items}
\begin{tabular}{@{}lcccccc@{}}
\toprule
System & Y04 & Y05 & Y06 & Y07 & Y08 & S10 \\
\midrule
Qwen2.5-VL-3B & 0.266 & 0.086 & 0.388 & 0.728 & 0.153 & 0.122 \\
Qwen2.5-VL-7B & 0.262 & 0.101 & 0.297 & 0.735 & 0.161 & 0.558 \\
InternVL2.5-4B & 0.266 & 0.117 & 0.416 & 0.731 & 0.156 & 0.503 \\
3B LoRA fields 10\% & 0.264 & 0.125 & 0.434 & 0.756 & 0.164 & 1.000 \\
7B LoRA fields 10\% & 0.269 & 0.116 & 0.418 & 0.731 & 0.158 & 0.991 \\
Qwen3-VL-Plus & 0.259 & 0.064 & 0.406 & 0.838 & 0.137 & 0.873 \\
GLM-4.5V & 0.254 & 0.101 & 0.384 & 0.757 & 0.156 & 0.855 \\
Step-1o & 0.247 & 0.110 & 0.405 & 0.749 & 0.139 & 0.915 \\
Doubao-Seed & 0.239 & 0.081 & 0.372 & 0.761 & 0.094 & 0.889 \\
HY-Vision & 0.242 & 0.063 & 0.377 & 0.754 & 0.140 & 0.966 \\
\bottomrule
\end{tabular}
\end{table}

The side column barely moves. Y04 runs from 0.239 on Doubao-Seed to 0.269 on the 7B adapter. Open zero-shot models, adapters, and closed models occupy the same narrow band around one quarter. A gain that was really a better reading of the frame would have shown up here. It does not. The side item is not where PhoneQC separates systems, and it is also not a saturated cell. There is room in it, and nobody in this table takes that room.

The screen column is the open cell. Y05 runs from 0.063 and 0.064 on HY-Vision and Qwen3-VL-Plus up to 0.125 on the 3B adapter. The adapter's own move, from 0.086 to 0.125, is small next to its blocker move. Closed models that score in the low 30s are among the weakest on the screen. A high Score100 on the answered set is compatible with missing the screen. That is why the task is not a solved detection problem, and why a detector metric alone would rank these systems differently from the form.

The back column moves a little and does not decide the rank. The 7B zero-shot model is the low end at 0.297. The 3B adapter is the high end at 0.434. InternVL2.5-4B is already at 0.416 before any adapter. The step from a strong zero-shot back score to the adapter is 0.018. The step from the 3B zero-shot back score of 0.388 to the adapter is 0.046. Both are real and both are small beside S10.

The residual other region is already high before adaptation. Y07 runs from 0.728 to 0.838, and the highest cell is Qwen3-VL-Plus at 0.838, a model whose coverage is 0.588. A dark cell in that column means the residual region is easy relative to the screen. It does not mean the form is complete. Severity, Y08, stays low for everyone, from 0.094 on Doubao-Seed to 0.164 on the 3B adapter. Doubao-Seed is the useful comparison inside that column. Its blocker item is 0.889 and its severity item is 0.094. A model can get the blocker and miss severity on the same orders. The two cells have to be reported separately, which is what the heatmap is for.

S10 is the cell that actually separates the 3B zero-shot model from field-only LoRA, 0.122 against 1.000. The 7B zero-shot model already reaches 0.558, and InternVL2.5-4B reaches 0.503, so some of the blocker is available without adaptation once the backbone is larger. The score only moves from 26.61 to 27.90 across that backbone step. S10 by itself does not set the rank. Closed models are high on S10, from 0.855 to 0.966, on the orders they answer. The 3B adapter is the open system that reaches 1.000 and still answers 311 orders. The training target wrote the staff fields and left the defect list empty. The item that moves is the staff blocker. The screen and the severity grade stay open, so the benchmark still has a hard appearance problem after the adapter is applied.

\paragraph{Listing-condition analysis.}
The cut-level mean treats a mint order and an inner-crack order as two draws from one task. The order identifier already carries a listing-condition token, and Table~\ref{tab:s-cond} scores the five tokens that cover 327 of the 335 orders. The other eight orders use a leftover token and are left out of this table. ``In cut'' counts every order with that token. The other columns are answered orders only, so a token with a smaller $n$ is a token that system did not always answer.

\begin{table}[t]
\centering\scriptsize
\setlength{\tabcolsep}{3.5pt}
\caption{Listing-condition tokens already on the order identifier.
``In cut'' counts every order with that token.
The other three columns are answered orders and the mean Score100 on those orders.
Five tokens cover 327 of 335 orders.
The remaining eight orders use a leftover token and are not in this table.}
\label{tab:s-cond}
\begin{tabular}{@{}lccccccc@{}}
\toprule
Token & In cut & \multicolumn{2}{c}{3B zero-shot} & \multicolumn{2}{c}{3B LoRA} & \multicolumn{2}{c}{Step-1o} \\
\cmidrule(lr){3-4}\cmidrule(lr){5-6}\cmidrule(lr){7-8}
 & & $n$ & Score & $n$ & Score & $n$ & Score \\
\midrule
Mint & 71 & 70 & 33.54 & 68 & 39.44 & 60 & 37.90 \\
Light wear & 128 & 124 & 26.57 & 125 & 33.48 & 101 & 32.27 \\
Heavy wear & 77 & 76 & 23.00 & 67 & 29.87 & 64 & 27.86 \\
Outer crack & 32 & 31 & 24.66 & 28 & 31.41 & 20 & 24.78 \\
Inner crack & 19 & 19 & 20.04 & 15 & 25.11 & 7 & 23.29 \\
\bottomrule
\end{tabular}
\end{table}

For the 3B zero-shot model the answered means fall from mint to inner crack, 33.54, 26.57, 23.00, 24.66, and 20.04. The fall is not monotone. Outer crack at 24.66 sits 1.66 above heavy wear at 23.00. The same inversion remains after adaptation. The 3B adapter scores 31.41 on outer crack and 29.87 on heavy wear. Step-1o goes the other way on those two tokens, 24.78 on outer crack against 27.86 on heavy wear, a gap of 3.08 in the opposite direction. The five tokens therefore do not form one order that every system obeys. PhoneQC reports the token because the spread is large, and it does not treat the token list as a condition ladder.

The spread that does show up on every row is mint against inner crack. The 3B zero-shot gap is 13.50, from 33.54 down to 20.04. The adapter gap is 14.33, from 39.44 down to 25.11. Step-1o, on the orders it answers, gaps by 14.61, from 37.90 down to 23.29. One average over the cut would hide a 13 to 15 point difference that is already written on the order identifier. The benchmark keeps the token so that a method cannot look strong by scoring only the mint bin.

The adapter is higher than the 3B zero-shot model in every bin, and the margin is not confined to mint. Mint rises by 5.90, light wear by 6.91, heavy wear by 6.87, outer crack by 6.75, and inner crack by 5.07. The gain on inner crack, where the zero-shot mean is 20.04, is the check against an easy-order artifact. The adapter still answers 15 of the 19 inner-crack orders. Step-1o's inner-crack mean of 23.29 rests on 7 orders, so that one cell is thin. The direction matches the other four tokens, where Step-1o's counts are 60, 101, 64, and 20. The closed model also finds mint easier than inner crack. It does so on fewer orders, which is the coverage fact from the previous section applied inside the token.

Answered counts also move a little across systems, and that movement is part of the table. The adapter answers 125 of 128 light-wear orders and only 15 of 19 inner-crack orders. Step-1o answers 60 of 71 mint orders and 7 of 19 inner-crack orders. The harder token is where answering thins out. A mean that ignored $n$ would let a 7-order cell look as solid as a 125-order cell. The table prints $n$ so the thin cell stays visible.

\section{Detector Sidecar and Robustness}
\label{sec:s-t2-rob}
The main paper already reports the compact T2 table and the robustness score/coverage table.
This section expands the same logs with grouped bars, staff-fill columns, and decision rates, so that a better box or a damaged upload cannot be mistaken for the T1 rank.

\begin{figure}[t]
\centering
\includegraphics[width=\linewidth]{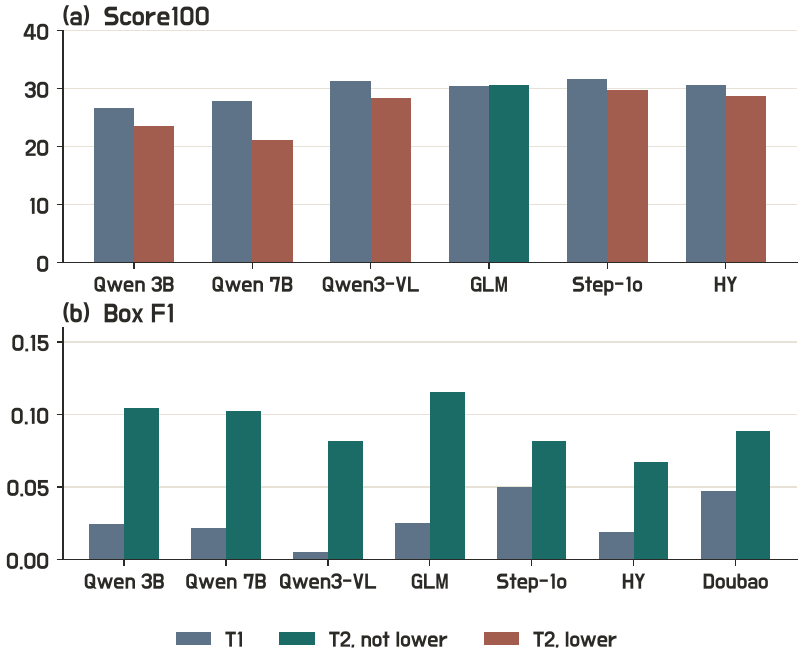}
\caption{Same backbone, T1 against T2, as grouped bars.
(a) Score100. (b) Box F1.
Box F1 rises for every backbone.
Score100 does not, except for a negligible rise on GLM-4.5V.
Doubao's T2 score is omitted because its T2 coverage is 0.17.}
\label{fig:s3}
\end{figure}

\paragraph{Detector sidecar on T2.}
T2 gives every backbone the same kind of extra input: predicted boxes and a short screen transcript are joined to the prompt, while reference boxes stay hidden.
If localization were the missing piece of the quote, Score100 would rise with box F1.
Figure~\ref{fig:s3} draws the two outcomes as grouped bars so the mismatch is visible without reading a delta off a table.

Panel (b) rises on every backbone. Panel (a) falls on every scored backbone except GLM-4.5V, whose rise is 0.14. The figure omits Doubao-Seed's T2 score because coverage on that run is 0.173. A bar at that coverage would be a mean on 58 orders, and the main paper therefore leaves it out of the comparison. The logged value is still in Table~\ref{tab:s-t2}, starred, so the omission is a ranking decision and not a missing log.

\begin{table}[t]
\centering\scriptsize
\setlength{\tabcolsep}{3.5pt}
\caption{T2 next to the T1 score of the same backbone.
$\Delta$ is the change in Score100.
n.a. means not applicable.
Doubao's T2 score is marked because coverage is 0.173.}
\label{tab:s-t2}
\begin{tabular}{@{}lccccccc@{}}
\toprule
Backbone & T1 & T2 & $\Delta$ & T2 cov. & T1 F1 & T2 F1 & T2 S-fill \\
\midrule
Qwen2.5-VL-3B & 26.61 & 23.50 & $-$3.11 & 0.997 & 0.024 & 0.105 & 0.257 \\
Qwen2.5-VL-7B & 27.90 & 21.13 & $-$6.77 & 0.988 & 0.022 & 0.103 & 0.182 \\
Qwen3-VL-Plus & 31.25 & 28.36 & $-$2.89 & 0.639 & 0.005 & 0.081 & 0.477 \\
GLM-4.5V & 30.44 & 30.58 & $+$0.14 & 0.994 & 0.025 & 0.116 & 0.461 \\
Step-1o & 31.65 & 29.72 & $-$1.93 & 0.982 & 0.050 & 0.082 & 0.453 \\
Doubao-Seed & 30.66 & 29.88$^\star$ & n.a. & 0.173 & 0.047 & 0.089 & 0.448 \\
HY-Vision & 30.64 & 28.70 & $-$1.94 & 0.734 & 0.019 & 0.067 & 0.493 \\
\bottomrule
\end{tabular}
\end{table}

For Qwen2.5-VL-3B, T2 lowers Score100 from 26.61 to 23.50 while box F1 rises from 0.024 to 0.105 and coverage rises from 0.976 to 0.997. Staff fill falls from 0.313 to 0.257. For the 7B model, Score100 falls from 27.90 to 21.13 and staff fill from 0.396 to 0.182, while box F1 rises from 0.022 to 0.103. Thus, better box scores do not guarantee a better form score; for these models the sidecar coincides with lower staff-component scores.

Several closed models answer more orders under T2 but still score lower. Qwen3-VL-Plus coverage rises from 0.588 to 0.639 while Score100 falls by 2.89; Step-1o coverage rises from 0.773 to 0.982 while Score100 falls by 1.93; HY-Vision coverage rises from 0.618 to 0.734 while Score100 falls by 1.94. Their T2 staff fill ranges from 0.448 to 0.493. These outcomes show that additional boxes and OCR do not consistently improve the composite form score.

GLM-4.5V is the exception, and the size of the exception is the point. The score rises from 30.44 to 30.58. Coverage also rises, from 0.824 to 0.994, so the gain of 0.14 is not produced by discarding hard orders. Box F1 rises from 0.025 to 0.116. Staff fill on T2 is 0.461, next to 0.433 on T1. The localization jump is large beside a score change of 0.14. That is the pattern the benchmark is designed to keep visible. Box F1 can improve while the form barely moves. Adding F1 into Score100 would have credited this row for a box that did not complete the quote.

Doubao-Seed is the coverage failure inside T2. The starred score is 29.88 and the coverage is 0.173, which is 58 answered orders. On this protocol the run sends 62.7\% of orders to human review. Box F1 still rises, from 0.047 to 0.089, and T2 staff fill is 0.448. The boxes got better on the orders that remained, and the system stopped answering most of the cut. The star is there so 29.88 cannot be read as a T2 result comparable to GLM's 30.58 at coverage 0.994.

The practical consequence is a rule for the leaderboard: a sidecar is allowed to raise box F1, but it is not allowed to take the image-only rank, because the rank is the form under a fixed input.
T1 and T2 are different inputs, and field-only LoRA stays on T1, where the images match the zero-shot images and the sidecar is absent.
The grouped bars are the evidence that those two tracks must stay apart.

The drop is easier to misread as a bad prompt than the staff-fill column allows.
On the open models, staff fill falls from 0.313 to 0.257 for the 3B model and from 0.396 to 0.182 for the 7B model, while box F1 rises to about 0.10. The T1 field-only adapter reaches a similar box F1 without a detector sidecar.
T2 reaches a similar F1 by showing boxes to a frozen model, yet the form gets worse: a better box and a filled form are not the same event.
GLM-4.5V is the most generous sidecar we log, a score change of 0.14 on a coverage that also rises from 0.824 to 0.994.
A deployment that wants the boxes can keep them, but it still cannot spend them as the image-only rank.

\begin{figure}[t]
\centering
\includegraphics[width=\linewidth]{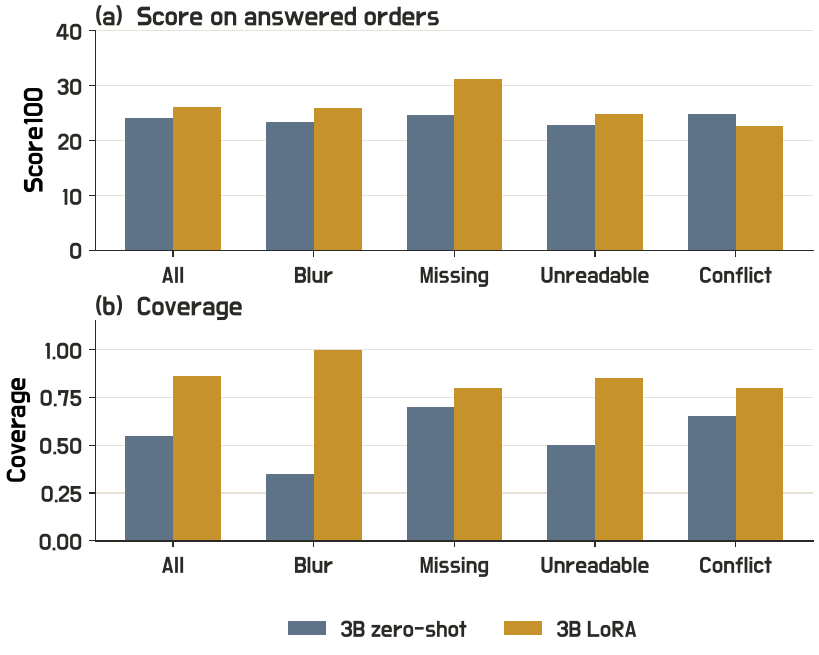}
\caption{Eighty corrupted orders.
(a) Score on the orders still answered. (b) Coverage.
Blur is the clearest coverage gap.
The answered score stays in the mid-20s for both models.}
\label{fig:s5}
\end{figure}

\paragraph{Robustness under visual corruptions.}
The comparisons above assume the six views and the screen are readable.
The robustness slice rescores 80 orders after blur, a missing view, an unreadable screenshot, or a modality conflict.
Figure~\ref{fig:s5} plots the answered score and the coverage, and Table~\ref{tab:s-rob} adds the decision rates the figure does not draw.
The 3B field-only adapter is the LoRA column.
These 80 orders are a stress test, not a second leaderboard, and the overall scores of 24.04 and 26.13 sit below the same models' scores on the full cut.

On the clean reading of panel (b), the adapter answers more of the damaged slice. Overall coverage is 0.86 against 0.55. Blur is the widest gap, 1.00 against 0.35. A missing view is 0.80 against 0.70. An unreadable shot is 0.85 against 0.50. A modality conflict is 0.80 against 0.65. The adapter's review rate in Table~\ref{tab:s-rob} is 0.000 on every row of this slice. When it does not answer, it abstains. The zero-shot model reviews 0.325 of the slice overall, 0.300 under blur, 0.250 on a missing view, 0.450 on an unreadable shot, and 0.300 on a conflict. Damage pushes the zero-shot model toward a person. The adapter keeps returning a form, or it abstains.

\begin{table}[t]
\centering\footnotesize
\setlength{\tabcolsep}{3.5pt}
\caption{Robustness on 80 orders, with the decision rates that Figure~\ref{fig:s5} does not draw.
Score100 is the mean on answered orders.}
\label{tab:s-rob}
\begin{tabular}{@{}lcccc@{}}
\toprule
 & \multicolumn{2}{c}{3B zero-shot} & \multicolumn{2}{c}{3B LoRA} \\
\cmidrule(lr){2-3}\cmidrule(lr){4-5}
Condition & Abstain & Review & Abstain & Review \\
\midrule
Overall & 0.125 & 0.325 & 0.138 & 0.000 \\
Blur & 0.350 & 0.300 & 0.000 & 0.000 \\
Missing view & 0.050 & 0.250 & 0.200 & 0.000 \\
Unreadable shot & 0.050 & 0.450 & 0.150 & 0.000 \\
Modality conflict & 0.050 & 0.300 & 0.200 & 0.000 \\
\bottomrule
\end{tabular}
\end{table}

The answered score does not move in lockstep with that coverage gain, which is why panel (a) is drawn separately. Overall, the adapter scores 26.13 against 24.04. Under blur the scores are 25.89 and 23.29, a gap of 2.60, while coverage goes from 0.35 to 1.00. The reliability claim on blur is the coverage. The score on the orders the zero-shot model still answers is already in the mid-20s, so blur does not reveal a hidden high score. On a missing view the score gap is larger, 31.25 against 24.66, and coverage moves only from 0.70 to 0.80. Here the adapter's advantage shows up in the form, not only in the decision to answer. On an unreadable shot the scores are 24.82 and 22.74, with coverage 0.85 against 0.50. The zero-shot review rate of 0.450 is the operational fact. Nearly half the unreadable orders are handed back.

The modality conflict is the row that keeps the claim honest. The zero-shot score is 24.76 and the adapter score is 22.69. Coverage still favors the adapter, 0.80 against 0.65. The adapter abstains on 0.200 and reviews none. The zero-shot model abstains on 0.050 and reviews 0.300. Answering more often under a contradictory prompt is not the same as scoring higher. The table keeps both columns so this row cannot be averaged away into the overall 26.13. The adapter is more willing to return a form when the upload is damaged. It is not higher on every corruption. The gap from either of these scores up to the oracle at 86.66 remains the whole point of the stress test. Damage does not create the gap. The gap is already there on clean T1.

A smaller note on a different 100-order slice asks whether the images are doing the work. Removing the images and leaving text makes the 3B zero-shot model abstain on 95\% of those orders, coverage 0.05. The mean on the few orders that still receive an answer is 26.58. That mean sits next to the image scores and must not be read as evidence that text works. Five percent of the slice produces it. An image-only run on the same slice scores 26.54 at coverage 0.98, next to the full prompt at 25.73, also at coverage 0.98. Box F1 is 0.041 for both the full prompt and the image-only run, and it is 0 for text only. The blocker item on this slice is 0.122 with the full prompt, 0.337 with images alone, and 0 with text alone. The images carry the answer the scorer can use. Text alone does not. A text leak would not pass this benchmark, because the model would abstain.

\section{Adaptation Ablations and Slices}
\begin{figure}[t]
\centering
\includegraphics[width=\linewidth]{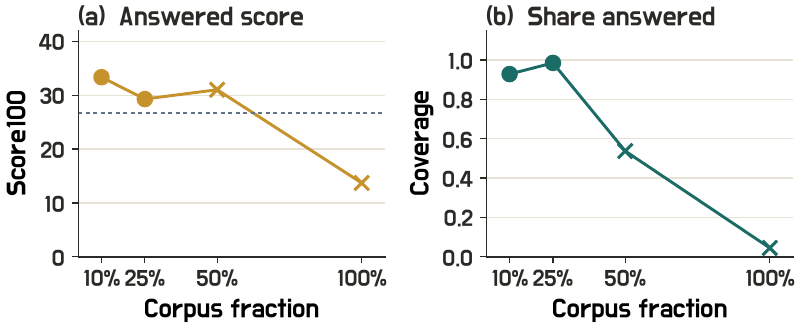}
\caption{Field-only LoRA on Qwen2.5-VL-3B as the corpus fraction grows.
(a) Score100. The dashed line is the 3B zero-shot score.
(b) Coverage.
Crosses are the 50\% and 100\% runs, where coverage falls below 0.9.}
\label{fig:s6}
\end{figure}

\paragraph{Corpus fraction and supervision target.}
The adapter result raises a training question that the leaderboard cannot answer by itself. How much of the isolated corpus, and which tokens inside the target, produce a score that still comes with a form. Figure~\ref{fig:s6} varies the fraction for field-only LoRA on the 3B backbone. The dashed line is the 3B zero-shot score of 26.61. Crosses mark the 50\% and 100\% runs, where coverage falls below 0.9. Each fraction is one training run, so the curve has no variance band. The shape is still the result. It is not a smooth climb.

At 10\%, the score is 33.33 with coverage 0.928; at 25\%, it is 29.27 with coverage 0.985. At 50\%, the score is 31.00 on 180 orders (coverage 0.537), staff fill is 0.494, appearance fill is 0.231, and box F1 is 0.061. Because coverage falls below 0.9, the 50\% mean is not a leaderboard result. Each fraction was run once, so the curve has no estimate of run-to-run variation.

At 100\%, the model answers 15 orders (coverage 0.045) and scores 13.70. Appearance fill is 0.088, staff fill is 0.251, and box F1 is 0.000. This single run under the fixed step budget scores below the zero-shot baseline of 26.61 and has very low coverage.

A shorter 50\% run, 200 steps, is the sharpest illustration of the same trap. The answered score is 35.01, the highest mean in Table~\ref{tab:s-extra}, and coverage is 0.376, which is 126 orders. The star is the entire content of that row. Without it, a reader would crown a run that refuses most of the cut. The benchmark's rule is that coverage below 0.9 removes the score from the board, even when the answered mean is the largest number in the column.

\begin{figure}[t]
\centering
\includegraphics[width=\linewidth]{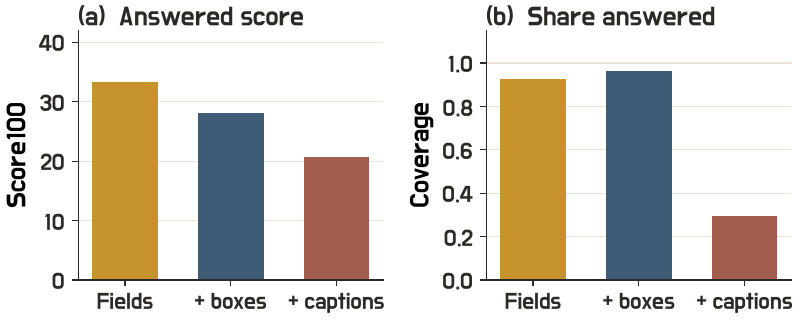}
\caption{What the 3B adapter is trained to emit, at 10\% of the corpus and the same step budget.
(a) Score100. (b) Coverage.
Supervising the staff fields raises the score and keeps coverage.
Adding boxes, then captions, does not.}
\label{fig:s7}
\end{figure}

Figure~\ref{fig:s7} holds the fraction at 10\% and changes the target, at the same step budget. The selected target writes the staff fields and closes the defect list as an empty array, so no box coordinate is a supervised token. The box ablation puts reference boxes into that list. The caption ablation appends caption sentences. All three are tested on T1. The images match the zero-shot images. No sidecar is handed over at test time.

Fields only scores 33.33 at coverage 0.928, with box F1 0.113. Writing boxes into the list lowers the score to 28.10 at coverage 0.964, with box F1 0.105. The box target does not even win the localization column against the empty defect list. Adding captions drops the score to 20.67 and the coverage to 0.296, which is 99 orders, with box F1 0.071. A heavier target makes the model stop answering. It is not a better target. The figure marks fields-only as the recipe that is kept because it is the setting that is both high and still answering.

\begin{table}[t]
\centering\footnotesize
\setlength{\tabcolsep}{3.5pt}
\caption{Logged runs that are not leaderboard points.
Stars mark coverage below 0.9.
MiniMax-M2 is a separate logged system, not a 3B training run.
The 100\% stable run answers, and its score stays below the 10\% field-only adapter.}
\label{tab:s-extra}
\begin{tabular}{@{}lccc@{}}
\toprule
Run & Score100 & Cov. & Answered \\
\midrule
Fields 50\% & 31.00$^\star$ & 0.537 & 180 \\
Fields 100\% & 13.70$^\star$ & 0.045 & 15 \\
Fields 50\%, 200 steps & 35.01$^\star$ & 0.376 & 126 \\
Fields + boxes, stable 10\% & 28.79$^\star$ & 0.490 & 164 \\
Fields + boxes, stable 25\% & 22.81$^\star$ & 0.821 & 275 \\
Fields + boxes, stable 50\% & 22.63$^\star$ & 0.687 & 230 \\
Fields + boxes, stable 100\% & 25.93 & 0.970 & 325 \\
Fields + boxes + captions & 20.67$^\star$ & 0.296 & 99 \\
MiniMax-M2 & 29.07$^\star$ & 0.164 & 55 \\
\bottomrule
\end{tabular}
\end{table}

Table~\ref{tab:s-extra} collects the other logged runs that a reader might otherwise promote. The four stable rows are additional schedules on a fields-and-boxes target. They do not climb past the short field-only adapter, and they do not climb in a straight line. At 10\% the stable score is 28.79 on 164 orders, coverage 0.490, so the star applies. At 25\% the score is 22.81 on 275 orders and coverage is 0.821, which is still below 0.9, so that row is starred as well. The score itself sits under the 3B zero-shot score of 26.61. At 50\% the score is 22.63 on 230 orders, coverage 0.687. At 100\% the run answers 325 orders, coverage 0.970, and scores 25.93. This is the clean negative among the stable rows. Coverage is high enough to read the score, and the score is below the zero-shot model and far below 33.33. More of the corpus, on that schedule, does not take the rank.

MiniMax-M2 is not one of those training runs. It is a separate logged system that answers 55 orders, coverage 0.164, with Score100 29.07. The mean sits near InternVL2.5-4B. The coverage does not. The star keeps 29.07 off the board for the same reason the 200-step mean of 35.01 stays off the board. An answered score is only a rank when the system also returns the form.

\begin{figure}[t]
\centering
\includegraphics[width=\linewidth]{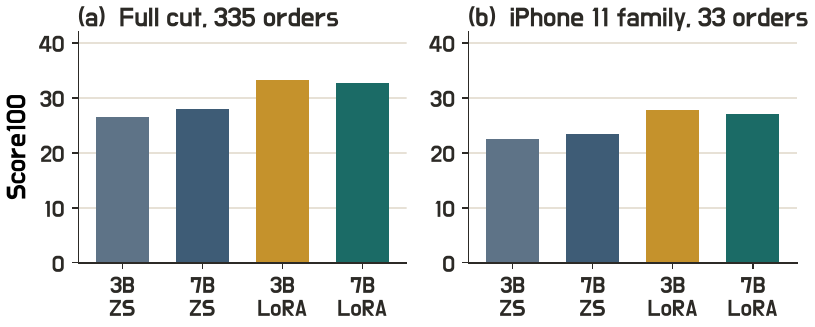}
\caption{The same four systems on two slices.
(a) The 335-order cut. (b) The 33-order iPhone~11 family.
Every bar drops from (a) to (b), and field-only LoRA stays above its zero-shot counterpart.}
\label{fig:s8}
\end{figure}

\paragraph{iPhone~11 family slice.}
Training uses other orders.
An audit on order identifiers and on image hashes finds no overlap with the 335.
Inside those 335, 33 orders belong to the iPhone~11 family.
Figure~\ref{fig:s8} draws the same four open systems on the full cut and on that family, asking whether the adapter's lead is an in-family memorization of the phones that dominate the cut, or whether the lead remains when the family changes.

Every bar drops. Table~\ref{tab:s-hold} prints the two means. The 3B zero-shot model falls from 26.61 to 22.54, a drop of 4.07. The 7B zero-shot model falls from 27.90 to 23.38, a drop of 4.52. The 3B adapter falls from 33.33 to 27.88, a drop of 5.45. The 7B adapter falls from 32.74 to 27.11, a drop of 5.63. The adapters drop by more points than their zero-shot counterparts. Part of the full-cut gain sits on phones outside this family. The ordering does not reverse. On the 33 orders the adapters still lead, 27.88 against 22.54, and 27.11 against 23.38.

\begin{table}[t]
\centering\footnotesize
\setlength{\tabcolsep}{3.5pt}
\caption{iPhone~11 family, 33 orders, from the holdout logs.
Full-cut scores are the T1 numbers in Table~\ref{tab:s-t1}.}
\label{tab:s-hold}
\begin{tabular}{@{}lcc@{}}
\toprule
System & Full & Holdout \\
\midrule
Qwen2.5-VL-3B & 26.61 & 22.54 \\
Qwen2.5-VL-7B & 27.90 & 23.38 \\
3B LoRA fields 10\% & 33.33 & 27.88 \\
7B LoRA fields 10\% & 32.74 & 27.11 \\
\bottomrule
\end{tabular}
\end{table}

Thirty-three orders are a slice, not a new benchmark. The drop says the family is harder than the cut mean, for every one of these four systems. The lead that remains says the field-only gain is not limited to the phones seen as the bulk of the cut. A holdout that had put the adapter under its zero-shot counterpart would have said the opposite. That is not the log. The absolute level, 27.88 on the better adapter, is still far from the oracle at 86.66. Holding the lead on a harder family does not close the form.

\begin{figure}[t]
\centering
\includegraphics[width=\linewidth]{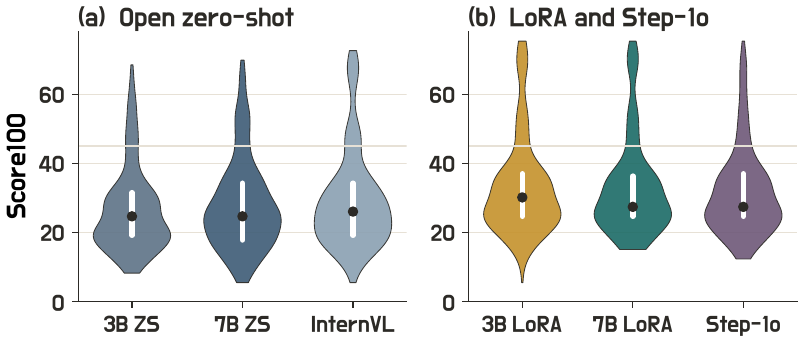}
\caption{Score100 on answered orders.
(a) Open zero-shot models. (b) Field-only LoRA and Step-1o.
The white bar is the interquartile range and the dot is the median.
Both panels share the scale.
The mass stays below the line at 45.}
\label{fig:s9}
\end{figure}

\paragraph{Score distribution.}
A mean of 30 can be misread as a cluster of orders near 30, or as a mix of solved orders and failed ones. Figure~\ref{fig:s9} draws the answered orders themselves. The white bar is the interquartile range and the dot is the median. Both panels share the scale. The mass stays below the line at 45.

Panel (a) is the open zero-shot models whose means are 26.61, 27.90, and 29.31. Panel (b) puts the field-only adapters, whose means are 33.33 and 32.74, next to Step-1o at 31.65. Sharing the scale is the point of the figure. If each panel were rescaled to its own maximum, the adapter would look as spread out as the zero-shot model even when both sit in the same low region of a 0 to 100 score. The line at 45 is there so the eye can see that the body of the distribution, not a handful of failed orders, is what holds the mean down. The leaderboard numbers in the high 20s and low 30s are inside that mass. They are not an average of solved orders and a few zeros.

The oracle at 86.66 is not shown because it is not a T1 system. The adapter shifts the answered-order distribution upward relative to the 3B zero-shot model, but scores remain below 45. The screen and severity cells remain low (Table~\ref{tab:s-items}), so the benchmark retains substantial headroom.

\begin{figure}[t]
\centering
\includegraphics[width=\linewidth]{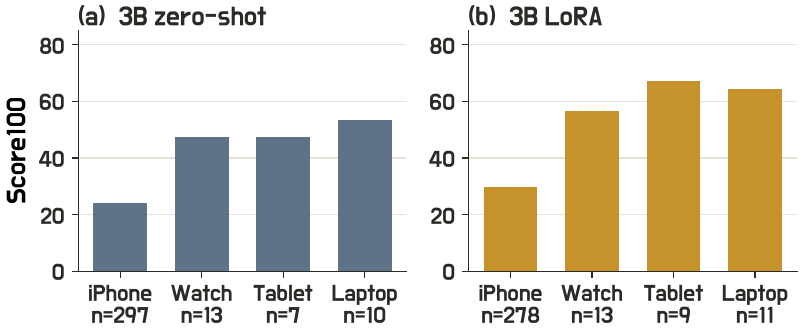}
\caption{Score100 by the device class written in the order identifier.
(a) 3B zero-shot. (b) 3B field-only LoRA.
The count under each label is the number of orders that system answered.
iPhones are the hard mass.
Watches, tablets, and laptops are few and score higher.}
\label{fig:s10}
\end{figure}

\paragraph{Device-class breakdown.}
The order identifier also names a device class. Figure~\ref{fig:s10} scores the classes on the orders each system answered. The count under each label is that answered size. Table~\ref{tab:s-device} is the same comparison in numbers. The zero-shot model leaves 8 orders unanswered and the adapter leaves 24, so the class counts are not the same set of orders.

iPhones are the mass, and they are the hard mass. The 3B zero-shot model answers 297 iPhones at a mean of 24.29, against 13 watches at 47.52, 7 tablets at 47.55, and 10 laptops at 53.42. Those four counts sum to 327, which is the answered total in Table~\ref{tab:s-decision}. About nine in ten answered orders are iPhones. The published zero-shot score of 26.61 sits 2.32 above the iPhone mean of 24.29 because the other 30 orders score in the high 40s and low 50s. A pooled mean lets a small non-phone group lift the number a reader will quote. Printing the count is what stops that lift from being mistaken for a property of the whole cut.

\begin{table}[t]
\centering\footnotesize
\setlength{\tabcolsep}{3.5pt}
\caption{Device class on the order identifier, answered orders only.
These are the bars in Figure~\ref{fig:s10}.
The zero-shot model leaves 8 orders unanswered, the adapter leaves 24, so the counts differ.}
\label{tab:s-device}
\begin{tabular}{@{}lcccc@{}}
\toprule
Class & \multicolumn{2}{c}{3B zero-shot} & \multicolumn{2}{c}{3B LoRA} \\
\cmidrule(lr){2-3}\cmidrule(lr){4-5}
 & $n$ & Score & $n$ & Score \\
\midrule
iPhone & 297 & 24.29 & 278 & 29.91 \\
Watch & 13 & 47.52 & 13 & 56.69 \\
Tablet & 7 & 47.55 & 9 & 67.12 \\
Laptop & 10 & 53.42 & 11 & 64.51 \\
\bottomrule
\end{tabular}
\end{table}

The adapter's iPhone mean is 29.91 on 278 orders, up from 24.29 on 297. The counts differ by 19 orders, so this is not a paired comparison on an identical set. It is the answered mean inside the class that dominates the cut, and it moves in the same direction as the overall score, from 26.61 to 33.33. Watches are the one class whose answered count matches, 13 against 13, and the mean rises from 47.52 to 56.69. Thirteen orders cannot carry a claim about watches. They can show that the adapter's direction is not an iPhone-only accident on the one class where $n$ is unchanged. Tablets go from 7 orders at 47.55 to 9 orders at 67.12, and laptops from 10 at 53.42 to 11 at 64.51. The sets changed. The means are high. The counts are too small to separate a device effect from the particular orders that happened to be answered.

The implication for the benchmark is about aggregation. A leaderboard that reported one mean and hid the class counts would let thirty non-phone orders pull an iPhone-heavy cut upward, and would let a 9-order tablet cell look like a solved device. PhoneQC prints $n$ under the bar. The hard problem in this cut is the iPhone mass at 24.29 before adaptation and 29.91 after it. The other classes are a warning about sample size, not a second result.

\section{Appraiser Agreement Study}
\label{sec:human}
The main paper leaves the human-reference check entirely in this supplement.
On 50 shared orders, 24 appraisers reach pairwise agreement 0.963 / 0.939 / 0.926 on model, storage, and grade (Fleiss $\kappa$ 0.961, 0.918, 0.896), defect-set F1 0.998 / 1.00 / 0.994, and majority-staff defect-set F1 0.94 / 0.82 / 0.90, with frame wear the softest cell.
On 138 stored view rectangles, pairwise matching F1 is 0.939 at IoU $0.5$, and 0.968 against the stored set.
These results provide a limited check on the staff reference for Score100; they do not establish uniform reliability across fields. Box F1 remains separate from the official rank.

\begin{figure}[t]
\centering
\includegraphics[width=\linewidth]{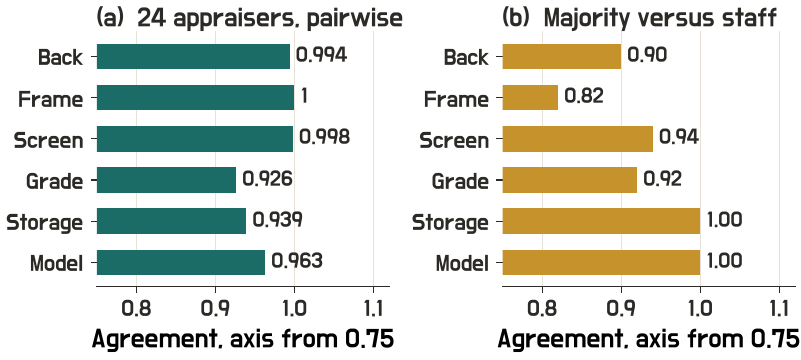}
\caption{Appraiser agreement on selected form fields.
Twenty-four appraisers filled the same 50 orders.
(a) Pairwise agreement. (b) Majority vote against the staff record.
The axis starts at 0.75.
Frame wear against staff is the soft cell, at 0.82.}
\label{fig:s12}
\end{figure}

\paragraph{Staff-form agreement.}
Score100 treats the staff form as the reference: the labels are that form, not scores written by a vision-language model, and that treatment is only as solid as the form itself.
Ethics consent forms for the orders were obtained.
Twenty-four professional appraisers then filled the same 50 orders after a shared naming guide, without drawing defect boxes in that pass.
Figure~\ref{fig:s12} draws pairwise agreement among the appraisers and the majority vote against the stored staff record, with the axis starting at 0.75 so that a cell at 0.82 is not visually identical to a cell at 1.00.

Table~\ref{tab:s-agree} reports the study in numbers. Pairwise agreement is 0.963 on model, 0.939 on storage, and 0.926 on grade, with Fleiss $\kappa$ of 0.961, 0.918, and 0.896. Grade is the least consistent of these fields. Defect classes agree among appraisers at set F1 0.998 on the screen, 1.000 on the frame, and 0.994 on the back. These figures indicate strong agreement on this 50-order subset; they do not imply that appraisers are interchangeable or that all scored fields have equally reliable references.

\begin{table}[t]
\centering\footnotesize
\setlength{\tabcolsep}{3.5pt}
\caption{Twenty-four appraisers on 50 orders.
Pairwise agreement and Fleiss $\kappa$ are among appraisers.
The staff column is majority vote against the staff record.
Frame wear is the soft cell.}
\label{tab:s-agree}
\begin{tabular}{@{}lccc@{}}
\toprule
Field & Pairwise & $\kappa$ & vs.\ staff \\
\midrule
Model & 0.963 & 0.961 & 1.00 \\
Storage & 0.939 & 0.918 & 1.00 \\
Grade & 0.926 & 0.896 & 0.92 \\
Screen defects & 0.998 & n.a. & 0.94 \\
Frame defects & 1.000 & n.a. & 0.82 \\
Back defects & 0.994 & n.a. & 0.90 \\
\bottomrule
\end{tabular}
\end{table}

The staff column is a different comparison, and it limits what a single score should be taken to mean. Majority vote matches the stored record on all 50 models and storages, on 46 of 50 grades, and at defect-set F1 0.94, 0.82, and 0.90 for screen, frame, and back. Frame wear at 0.82 is the least-aligned cell. Appraisers agree with one another on frame defects at 1.000 but agree less closely with the stored record. This comparison cannot identify which source is correct; it shows that the stored label is less settled for frame wear. Table~\ref{tab:s-items} therefore keeps defect regions separate. Severity, color, battery health, lock, and account status are unanimous among appraisers on this subset, but the stored form leaves most of these fields empty, so they are not scored against staff. Agreement among annotators alone does not validate a reference label.

\paragraph{View-rectangle agreement.}
The same appraisers also redrew stored view rectangles. Those rectangles mark views and screenshot panels. They are not scratch boxes. A defect-only cut of the rectangles is empty, which is the direct check. The rectangles that do exist sit on 124 of 642 images, 138 rectangles in the stored set. Pairwise matching at an overlap of at least one half, on those 124 images, has F1 0.939 and mean overlap 0.971. Against the stored rectangles the F1 is 0.968 and the mean overlap is 0.958. If the 642 images were all counted, pairwise F1 would read 0.988, because most images contain no rectangle and two appraisers who both draw nothing would match. Empty images are left out so that agreement is not inflated by shared blanks. The number the benchmark quotes is 0.939 on the images that contain a rectangle, and 0.968 against the stored set. Localization of views is stable. Localization of scratches is not what this pass measured, and box F1 in the leaderboard should not be described as a scratch-box score.

\paragraph{What the study licenses.}
The study supports a qualified reading of the benchmark results. Agreement is high for model and storage and among appraisers on the measured defect classes, while agreement with the stored record is lower for frame wear. The benchmark reports this softer cell separately; appearance fill also remains near 0.25 in every learned row. The study does not validate every field in the stored form or show that inspection is solved.

\section{Sensitivity without S10 and Common-Answered Subsets}
\label{sec:s-sens}
The main paper already states that every reference blocker list is empty, so S10 awards a perfect score to an empty prediction.
This section adds the evaluation volume asked for by that disclosure: Score100 recomputed with the blocker item removed from both numerator and denominator, staff items restricted to fields that can change, and paired means on orders answered by several systems together.

\paragraph{Score100 without S10.}
Table~\ref{tab:s-nos10} repeats the T1 answered means after dropping S10.
Open zero-shot Score100 without S10 is 27.44 / 26.28 / 28.09 for 3B, 7B, and InternVL.
Field-only LoRA at 10\% falls from 33.33 to 29.47, and the 7B adapter from 32.74 to 28.89.
The open ranking still places the 3B adapter first on this cut, ahead of InternVL at 28.09 and the 3B zero-shot model at 27.44.
Closed answered means without S10 sit between 26.82 and 28.18, still below the adapter on the orders each system answers, but those sets remain different.

Composite S-fill without S10 is nearly flat for the 3B zero-shot and 10\% adapter rows (0.36 against 0.36), so the large S-fill lift in the main table is not evidence of broad staff-field extraction once the empty blocker is removed.
Model (S01) moves from 0.84 to 0.88 and storage (S02) from 0.92 to 0.92 on those same answered sets; battery/lock items that stay empty in the staff record contribute little.
We therefore limit ``staff-field improvement'' claims on this cut to Score100 without S10 and to identity fields that actually move, not to composite S-fill driven by S10.

\begin{table}[t]
\centering\scriptsize
\setlength{\tabcolsep}{3.5pt}
\caption{T1 answered means with and without the blocker item S10.
$n$ is the answered count used for both columns.
S01 / S02 are model and storage item scores on the same answered set.}
\label{tab:s-nos10}
\begin{tabular}{@{}lcccccc@{}}
\toprule
System & $n$ & Score100 & w/o S10 & S-fill & S-fill w/o S10 & S01 / S02 \\
\midrule
Qwen2.5-VL-3B & 327 & 26.61 & 27.44 & 0.31 & 0.36 & 0.84 / 0.92 \\
Qwen2.5-VL-7B & 317 & 27.90 & 26.28 & 0.40 & 0.36 & 0.84 / 0.93 \\
InternVL2.5-4B & 334 & 29.31 & 28.09 & 0.38 & 0.35 & 0.82 / 0.90 \\
3B LoRA fields 10\% & 311 & 33.33 & 29.47 & 0.48 & 0.36 & 0.88 / 0.92 \\
7B LoRA fields 10\% & 335 & 32.74 & 28.89 & 0.48 & 0.36 & 0.91 / 0.94 \\
3B LoRA fields 25\% & 330 & 29.27 & 25.21 & 0.39 & 0.26 & 0.60 / 0.69 \\
Qwen3-VL-Plus & 197 & 31.25 & 28.00 & 0.45 & 0.36 & 0.83 / 0.91 \\
GLM-4.5V & 276 & 30.44 & 27.24 & 0.43 & 0.34 & 0.89 / 0.69 \\
Step-1o & 259 & 31.65 & 28.18 & 0.46 & 0.36 & 0.83 / 0.89 \\
Doubao-Seed & 234 & 30.66 & 27.28 & 0.47 & 0.37 & 0.89 / 0.94 \\
HY-Vision & 207 & 30.64 & 26.82 & 0.47 & 0.35 & 0.81 / 0.91 \\
\bottomrule
\end{tabular}
\end{table}

\paragraph{Common-answered subsets.}
Table~\ref{tab:s-common} restricts the mean to orders answered by every listed system.
On the 202-order intersection of 3B zero-shot, InternVL, 3B LoRA 10\%, Step-1o, and GLM-4.5V, LoRA scores 33.36 against Step 32.45 and InternVL 29.70; without S10 the same order is 29.49 / 29.02 / 28.61.
That paired check supports treating the high-coverage open lead as not an artifact of skipping closed-model leftovers alone.
On a tighter 75-order intersection that also forces Qwen3, Doubao, and HY-Vision to answer, LoRA scores 32.05 against Doubao 31.80 with S10 included, but without S10 Doubao is slightly ahead (28.54 against 28.12).
We therefore treat the 75-order cell as exploratory: $n$ is small, coverage selection is severe, and it is not used to set the image-only rank.
The ranking claim we retain is the full-cut open comparison with and without S10, together with the 202-order common subset where LoRA remains first under both scorings.

\begin{table}[t]
\centering\scriptsize
\setlength{\tabcolsep}{3.5pt}
\caption{Paired Score100 on common-answered orders.
Top block: 202 orders answered by 3B zs, InternVL, 3B LoRA 10\%, Step, and GLM.
Bottom block: 75 orders that also include Qwen3, Doubao, and HY.}
\label{tab:s-common}
\begin{tabular}{@{}lccc@{}}
\toprule
System & $n$ & Score100 & w/o S10 \\
\midrule
\multicolumn{4}{@{}l@{}}{(A) Common $n{=}202$} \\
3B zero-shot & 202 & 27.27 & 28.13 \\
InternVL2.5-4B & 202 & 29.70 & 28.61 \\
3B LoRA fields 10\% & 202 & 33.36 & 29.49 \\
Step-1o & 202 & 32.45 & 29.02 \\
GLM-4.5V & 202 & 30.91 & 27.60 \\
\midrule
\multicolumn{4}{@{}l@{}}{(B) Common $n{=}75$ (exploratory)} \\
3B zero-shot & 75 & 26.10 & 27.30 \\
3B LoRA fields 10\% & 75 & 32.05 & 28.12 \\
Qwen3-VL-Plus & 75 & 30.23 & 27.04 \\
GLM-4.5V & 75 & 29.67 & 26.45 \\
Step-1o & 75 & 31.73 & 28.39 \\
Doubao-Seed & 75 & 31.80 & 28.54 \\
HY-Vision & 75 & 31.45 & 27.56 \\
\bottomrule
\end{tabular}
\end{table}

\section{Implementation Notes}
\paragraph{Training and decoding.}
The selected adapter is the 3B field-only run on Qwen2.5-VL-3B-Instruct.
LoRA is applied to the query, key, value, and output projections with rank 8, $\alpha{=}16$, dropout 0.05, and no bias, while the vision encoder stays frozen.
Training uses the Hugging Face Trainer around that adapter with AdamW, learning rate $2\times10^{-4}$, no warmup schedule, batch size 1, four gradient-accumulation steps, bfloat16, and gradient checkpointing.
The run lasts 400 steps, recorded as about one epoch, and the script default keeps two images in each training instance on one local GeForce RTX 4090.
Closed models are not trained: they are called as APIs.

Decoding is greedy for every open model.
Qwen2.5-VL zero-shot and InternVL2.5-4B stop at 512 new tokens, while LoRA evaluation stops at 1024 new tokens.
Closed APIs are called at temperature 0 with a budget of 4096 tokens.
For GLM-4.5V and Doubao-Seed the request disables the thinking trace, because that trace was consuming the token budget and returning an empty form.
The same prompt, image contract, and parser are used for every system inside one protocol; T2 adds the detector sidecar and does not change the decoder.

\paragraph{Reproducibility checklist.}
Train and test orders are isolated by order identifier and by SHA-256 of image bytes; the release CSV stores only the hash, not the raw id.
The three-image pack is built by a fixed priority over exterior slots (front, back, side/bottom), resized to long side 448, without using staff labels or model scores to choose views; every system in one protocol sees that same pack.
Open checkpoints are Qwen2.5-VL-Instruct and InternVL2.5-4B as named in the main tables; closed scores come from the named API endpoints called greedily (temperature 0) with thinking traces disabled where required to avoid empty forms.
Unparsable JSON or missing required keys count as no answer or as empty fields under the published scorer; unparsable boxes count as false positives in box F1.
The package releases aggregate leaderboards, the 50-order agreement table, hashed order rows, and privacy-checked sample views; the full photograph corpus and personal identifiers stay withheld, so exact API calendar dates and private image paths cannot be redistributed with the anonymous package.

\paragraph{Trivial constant predictors.}
Score100 can look moderate when a large part of the form is shared by many orders, so the check is a constant form that never sees the images.
Two constants are scored with the same scorer as the leaderboard.
The empty form answers every order and fills nothing.
The majority form answers every order with the most common product model on this cut (iPhone 11), the most common storage (256GB), and a scene flag of true; that flag is true on all 335 orders, so the scene bit is free, and defects stay empty.
Both constants are fit on this cut, so they are an upper bound on guessing from the label margin, not a system that could be trained on a held-out pool.
Staff-oracle and detector-oracle ceilings remain privileged references, not image-only baselines.

\begin{table}[t]
\centering\footnotesize
\setlength{\tabcolsep}{3.5pt}
\caption{Constants scored with the same scorer as the leaderboard.
The empty form and the majority form do not see images.
The two learned rows are the logged T1 numbers, copied here for the comparison.}
\label{tab:trivial}
\begin{tabular}{@{}lccccc@{}}
\toprule
System & Score100 & Cov. & Y-fill & S-fill & Box F1 \\
\midrule
Empty form & 21.19 & 1.000 & 0.225 & 0.182 & 0.104 \\
Majority form & 28.13 & 1.000 & 0.264 & 0.321 & 0.104 \\
Qwen2.5-VL-3B & 26.61 & 0.976 & 0.246 & 0.313 & 0.024 \\
3B LoRA fields 10\% & 33.33 & 0.928 & 0.270 & 0.479 & 0.113 \\
\bottomrule
\end{tabular}
\end{table}

The empty form scores 21.19. The majority form scores 28.13. The 3B zero-shot model scores 26.61, which is above the empty form and below the majority form. Looking at the images is not, by itself, ahead of guessing the common model and the common storage. The 3B adapter scores 33.33, which is 5.20 above the majority form.
Part of that margin is S10: constants also score 1.000 on the empty blocker list, so the gap that remains informative after removing S10 is the open ranking in Table~\ref{tab:s-nos10}, not the raw S-fill comparison against 0.321.
Appearance fill does not carry the result: the constant already reaches 0.264, and the adapter reaches 0.270.

Box F1 on both constants is 0.104, above the zero-shot model's 0.024. An empty defect list matches the orders whose reference list is empty, and a predicted box that does not match is a false positive. A higher box F1 can therefore be the result of predicting no box. The same reading applies to the blocker item.

\section{Scope and Reporting}
\paragraph{Reporting guidelines.}
Score100 is averaged over answered orders. Report it with coverage or the answered count because different systems answer different subsets. For example, Step-1o scores 31.65 on 259 orders (coverage 0.773), while the 3B adapter scores 33.33 on 311 (0.928). These means are not a paired comparison. Other logged runs make the same point: the 200-step adapter scores 35.01 on 126 orders, and MiniMax-M2 scores 29.07 on 55; neither meets the 0.9 coverage threshold used for the leaderboard.

Condition and device breakdowns are descriptive. Five listing tokens cover 327 orders, and eight are excluded from token means. Token scores do not form a monotonic scale from mint to cracked: for the 3B zero-shot model, outer-crack orders score 24.66, above heavy-wear orders at 23.00. Device-class means also use small, system-specific answered sets; the 3B zero-shot and adapter tablet means, for example, use 7 and 9 orders, respectively.

S10 requires special care. Every reference blocker list is empty, so an empty prediction scores 1.000 even without image understanding. The 3B adapter's S10 increase from 0.122 to 1.000 contributes to its S-fill increase from 0.313 to 0.479 and to Score100. We therefore report the measured scores but do not interpret this change as blocker recognition or general staff-field extraction. The empty-list design also explains why constant predictors score well on S10.

The oracle score of 86.66 is a privileged upper bound, not a T1 result or a maximum possible score; its staff fill is 0.590. The best image-only result, 33.33 at coverage 0.928, remains far below it. The appearance component stays between 0.229 and 0.270 across learned rows, and screen and severity scores remain low. Box F1 is reported separately from Score100.

\paragraph{Scope of the present cut.}
This cut is dominated by iPhones: 297 of the 3B zero-shot model's 327 answered orders are iPhones. Other device classes have only 7--13 answered orders each, so their means are exploratory. The 33 iPhone~11 orders provide one within-family check; all four open systems score lower on this subset than on the full cut, although the adapters remain above their zero-shot counterparts. These data do not support broad claims about Android devices or other trade-in markets.

The 10\% field-only run is the strongest high-coverage configuration among the schedules tested, not evidence that less training data is generally better. Training fractions were compared under a fixed 400-step budget, and each fraction was run once; they are not epoch-matched or multi-seed estimates. The results therefore describe these logged runs and should be interpreted as such.

Each prompt uses at most three images from an order that may contain more views. Selection follows a fixed exterior-slot priority shared by every system in a protocol and does not use staff labels; the evaluation manifest should still record the chosen filenames for audit. The 335-order cut and its current annotations also limit what can be said about rare blockers and device families.

\paragraph{Summary of findings.}
Across the analyses, the 3B adapter raises Score100 from 26.61 to 33.33 at coverage 0.928 on this cut, while appearance fill remains near one quarter. Because the blocker references are all empty, composite S-fill is not used as evidence of broad staff extraction; Score100 without S10 and the 202-order common subset still place the adapter first among the open systems logged here. Adding boxes or captions to the training target lowers the logged Score100 under the fixed step budget, and the logged detector sidecar improves box F1 but does not consistently improve the form score. The human-reference study finds strong pairwise agreement on identity fields, with weaker agreement against the stored record on frame defects. These findings support PhoneQC as a benchmark observation under declared inputs; they do not show that appraisal is replaced or that the result transfers beyond this split.

\end{document}